\documentclass[10pt,twocolumn,letterpaper]{article}

\usepackage{cvpr}              %
\usepackage{graphicx}
\usepackage{amsmath}
\usepackage{amssymb}
\usepackage{booktabs}
\usepackage{multirow}
\usepackage[utf8]{inputenc} %
\usepackage[T1]{fontenc}    %
\usepackage[accsupp]{axessibility}

\usepackage{amsmath,amsfonts,bm}

\def\eqref#1{equation~\ref{#1}}

\def\1{\bm{1}}

\DeclareMathAlphabet{\mathsfit}{\encodingdefault}{\sfdefault}{m}{sl}
\SetMathAlphabet{\mathsfit}{bold}{\encodingdefault}{\sfdefault}{bx}{n}

\DeclareMathOperator*{\argmax}{arg\,max}

\usepackage{color,xcolor}
\usepackage{epsfig}
\usepackage{adjustbox}
\usepackage{array}
\usepackage{booktabs}
\usepackage{colortbl}
\usepackage{wrapfig}
\usepackage{hhline}
\usepackage{multirow}

\usepackage{paralist}
\usepackage{tabularx}

\usepackage{amsmath,amsfonts,amssymb}
\usepackage{bm}
\usepackage{nicefrac}
\usepackage{microtype}

\usepackage{changepage}
\usepackage{extramarks}
\usepackage{fancyhdr}
\usepackage{lastpage}
\usepackage{setspace}
\usepackage{soul}
\usepackage{xspace}

\usepackage{tabularx}

\usepackage{url}

\usepackage{enumerate}
\usepackage{enumitem}  %

\usepackage{makecell}

\usepackage{pifont} %
\usepackage[ruled,vlined]{algorithm2e}
\usepackage{tabularx} %
\newcolumntype{L}[1]{>{\raggedright\let\newline\\\arraybackslash\hspace{0pt}}m{#1}}
\newcolumntype{C}[1]{>{\centering\let\newline\\\arraybackslash\hspace{0pt}}m{#1}}
\newcolumntype{R}[1]{>{\raggedleft\let\newline\\\arraybackslash\hspace{0pt}}m{#1}}

\newcommand{\sect}[1]{Section~\ref{#1}}

\newcommand{\fig}[1]{Figure~\ref{#1}}

\newcommand{\ignore}[1]{}

\makeatletter
\DeclareRobustCommand\onedot{\futurelet\@let@token\@onedot}
\def\@onedot{\ifx\@let@token.\else.\null\fi\xspace}

\def\eg{e.g\onedot} 
\def\ie{i.e\onedot}

\def\etal{et al\onedot}

\makeatother

\definecolor{MyDarkBlue}{rgb}{0,0.08,1}
\definecolor{MyDarkGreen}{rgb}{0.02,0.6,0.02}
\definecolor{MyDarkRed}{rgb}{0.8,0.02,0.02}
\definecolor{MyDarkOrange}{rgb}{0.40,0.2,0.02}
\definecolor{MyPurple}{RGB}{111,0,255}
\definecolor{MyRed}{rgb}{1.0,0.0,0.0}
\definecolor{MyGold}{rgb}{0.75,0.6,0.12}
\definecolor{MyDarkgray}{rgb}{0.66, 0.66, 0.66}

\definecolor{bittersweet}{rgb}{1.0, 0.44, 0.37}
\definecolor{ballblue}{rgb}{0.13, 0.67, 0.8}
\definecolor{amethyst}{rgb}{0.6, 0.4, 0.8}
\definecolor{blue-violet}{rgb}{0.54, 0.17, 0.89}
\definecolor{brightpink}{rgb}{1.0, 0.0, 0.5}

\newcommand{\xhdr}[1]{\noindent\textbf{#1}} 
\usepackage[pagebackref,breaklinks,colorlinks]{hyperref}

\usepackage[capitalize]{cleveref}
\crefname{section}{Sec.}{Secs.}
\Crefname{section}{Section}{Sections}
\Crefname{table}{Table}{Tables}
\crefname{table}{Tab.}{Tabs.}

\newcommand{\model}{Programmatic Motion Concepts\xspace}
\newcommand{\modelshort}{PMC\xspace}

\def\cvprPaperID{2102} %
\def\confName{CVPR}
\def\confYear{2022}

\begin{document}

\title{Programmatic Concept Learning for Human Motion Description and Synthesis}

\author{Sumith Kulal\thanks{\xspace and $^\dagger$ indicate equal contribution. Project page: \url{https://sumith1896.github.io/motion-concepts}}\\
Stanford University\\
\and
Jiayuan Mao\footnotemark[1]\\
MIT \\
\and
Alex Aiken$^\dagger$\\
Stanford University\\
\and
Jiajun Wu$^\dagger$\\
Stanford University\\
}

\maketitle
\pagestyle{empty}
\thispagestyle{empty}

\begin{abstract}

We introduce \model, a hierarchical motion representation for human actions that captures both low-level motion and high-level description as motion concepts. This representation enables human motion description, interactive editing, and controlled synthesis of novel video sequences within a single framework. We present an architecture that learns this concept representation from paired video and action sequences in a semi-supervised manner. The compactness of our representation also allows us to present a low-resource training recipe for data-efficient learning. By outperforming established baselines, especially in the small data regime, we demonstrate the efficiency and effectiveness of our framework for multiple  applications.

\end{abstract}

\section{Introduction}
\label{sec:intro}

The advent of new datasets and progress in machine learning has pushed the boundaries of several video analysis tasks. In particular, there has been tremendous progress on description tasks such as action recognition~\cite{niebles2007hierarchical, soomro2012ucf101, kuehne2011hmdb, karpathy2014large, simonyan2014two} and localization~\cite{shou2016temporal,zeng2019graph, carreira2017quo, feichtenhofer2019slowfast, feichtenhofer2020x3d, wang2018non}, and on synthesis tasks such as human motion synthesis~\cite{habibie2017recurrent, guo2020action2motion, petrovich2021action, shlizerman2018audio, lin2018generating,ahuja2019language2pose, ghosh2021synthesis} and video synthesis~\cite{chan2019everybody, tulyakov2018mocogan, siarohin2019first, ren2020human}. However, most of these models focus solely on their respective tasks of video description or synthesis. We posit that these tasks are better learned jointly. We present a single framework for human motion description (recognizing and temporally localizing individual actions in the video), synthesis (generating videos from abstract descriptions) and editing (adding or removing actions and other fine-grained manipulations).

\begin{figure*}[t]
  \centering
  \vspace{-1.5em}
   \includegraphics[width=\linewidth]{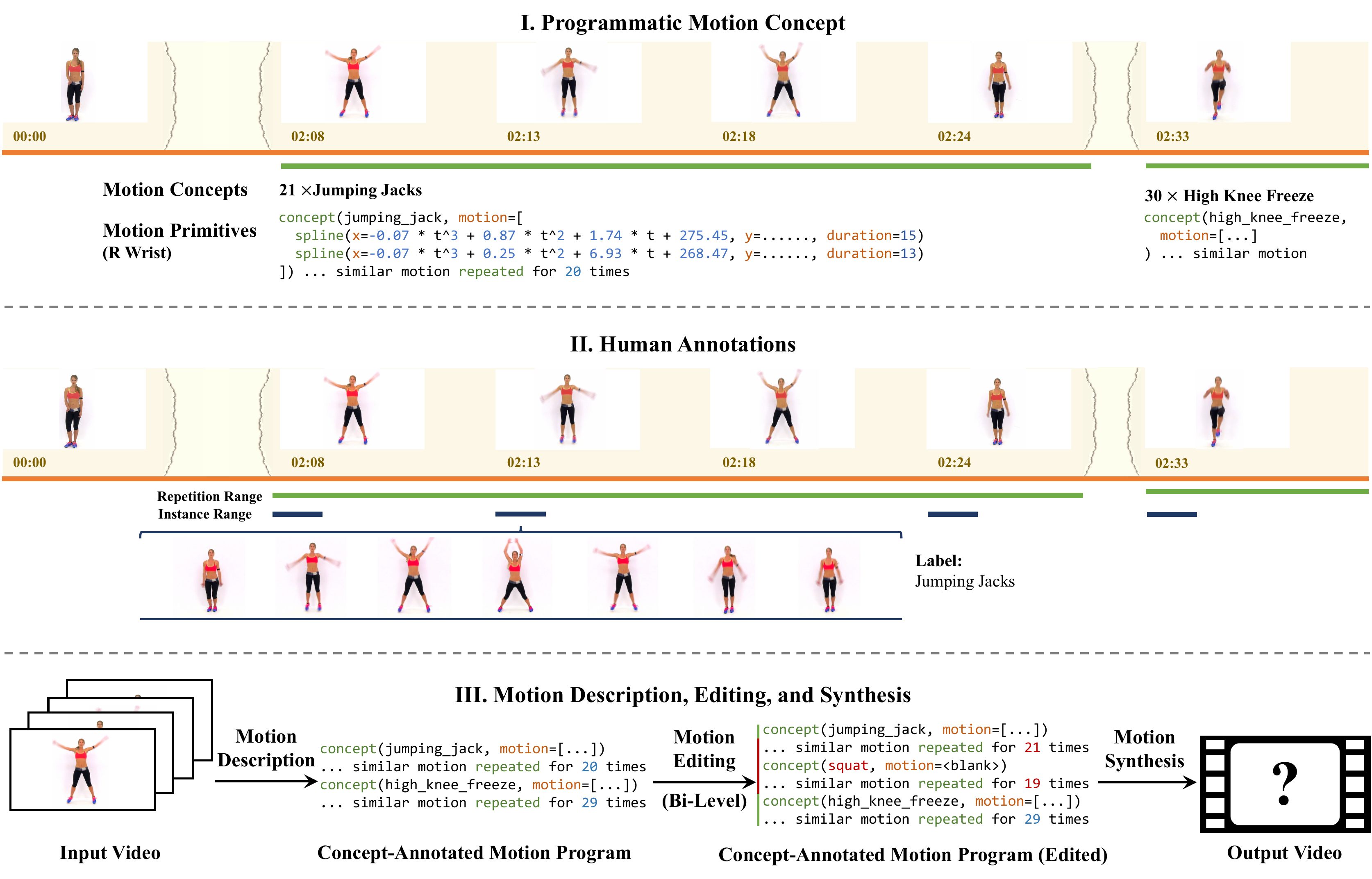}
   \vspace{-2em}
   \caption{(I) We present a hierarchical human motion description framework. Each video is represented as a sequence of {\it motion concepts} and each {\it motion concept} is further grounded as a sequence of {\it motion primitives}. (II) Motion concepts can be learned from very small amount of human annotation: human annotators label a repetition range of a motion concept and just three instance ranges with in this group of repetitions. (III) Motion concept supports interactive editing and video synthesis. Human editors can flexibly edit a human motion video at both the concept level or the primitive level. We use neural generative models to render the output video.}
   \vspace{-0.5em}
   \label{fig:teaser-figure}
\end{figure*}

 Our key insight is a hierarchical motion representation, {\it \model} (\modelshort). Beginning with standard low-level input (sequences of keypoints), we learn a representation of the distribution of high-level motion concepts (such as jumping jacks) first from a few examples and then from longer sequences of repetitions of the concept that we automatically segment. More specifically, for each motion concept and each body keypoint, we learn a distribution of time-space trajectories expressed as cubic splines. The fact that we explicitly learn a distribution of motions at the level of the entire motion concept is what makes it possible to perform recognition and localization well, since our methods leverage information about the motion of individual keypoints over the entire span of the motion concept.  Similarly, sampling from these distributions allows us to directly synthesize natural-looking instances of the motion concept.

\modelshort  is based on prior work on hierarchical motion understanding with programmatic primitive-based representations~\cite{motion2prog}. Illustrated in Figure~\ref{fig:teaser-figure}(I), a human motion sequence is represented as a sequence of {\it motion primitives}, which are compact, and human-interpretable, parametric curves. We further represent {\it motion concepts} as groupings of consecutive motion primitives that are named by humans, such as {\it jumping jack} and {\it squat} in workout videos, as shown in Figure~\ref{fig:teaser-figure}(I). This representation choice has access to both low-level pose sequences and high-level description sequences as motion concepts. 

We want to enable easy and efficient learning of our representation to quickly adapt this framework to different domains. Hence, we propose to learn this representation from {\it weakly-annotated} data. Our input data contains each concept annotated with only a few mouse clicks by the human annotator. For example, the annotation of each concept in our dataset consists of weak labels from $<15$ workout videos. Illustrated in Figure~\ref{fig:teaser-figure}(II), in each video, the human annotator provides a start and end point for a video segment that contains repetitions of jumping jacks as well as the start and
end points of three individual jumping jack instances. Our method learns two models from this data, a {\it recognition} model that can be applied on any human motion sequences to detect occurrences of jumping jacks and a {\it generative} model that can synthesize jumping jack motions. 

The key ideas behind our data efficient learning algorithm are three-fold. First, instead of using human skeleton sequences as input, our model operates on a representation of motion as time-space curves, which is a strong but generically applicable inductive bias for videos of human motion. Second, our model exploits the repetition of motion concepts in a single video. By annotating the beginning and the end of the repetition and a few occurrences of the action, our model automatically extrapolates to all instances in the repetition span. Third, the recognition and generation models are trained jointly, aiming at localizing individual occurrences of actions in the training data and learning the corresponding distribution of motion sequences, which significantly improves the quality of our motion synthesis model.
\model supports multiple human video analysis tasks, including recognition, localization, synthesis, and editing. Illustrated in \fig{fig:teaser-figure}(III), given the input motion sequence, our model produces a human-interpretable label sequence of  motion concepts. Each label is localized to a time span of the input. Users can tune the parameters for different motion primitives and even edit the label sequence itself to synthesize new motion sequences. Combined with techniques for human skeleton detection and skeleton-to-video synthesis, our data-efficient methods can be used to construct a full, practical pipeline for video-to-concept and concept-to-video workflows.

To summarize, our contributions are: 
\begin{itemize}[noitemsep,topsep=0pt,parsep=0pt,partopsep=0pt,leftmargin=2em]
\setlength\itemsep{-0.1em}
\item We present a novel hierarchical representation of human motion that jointly supports description, synthesis, and editing of human motion videos.
\item We present an data-efficient learning algorithm that leverages a motion primitive-based representation and repetitive structures in videos.
\item Finally, we demonstrate the efficiency and effectiveness of our concept learning framework on three downstream tasks: motion description, action-conditioned motion synthesis, and controlled motion and video synthesis. We also present qualitative results on interactive editing. 
\end{itemize}
\vspace{-0.5em}
\section{Related Work}
\vspace{-0.5em}
\label{sec:related-works}
\paragraph{Action recognition and localization.} Large datasets~\cite{soomro2012ucf101, kuehne2011hmdb, carreira2017quo} have led to deep learning architectures performing data-driven action recognition without hand-designed features~\cite{simonyan2014two, donahue2015long, tran2015learning, wang2015action}. Recent works also tackle the closely related problem of localization of actions~\cite{caba2015activitynet, idrees2017thumos, yeung2016end, zhao2017temporal, chao2018rethinking} that predict the temporal bounds of actions. Current state-of-the-art approaches for recognition~\cite{liu2021video, ryoo2021tokenlearner, arnab2021vivit} and localization~\cite{wang2021proposal} involve variants of the Transformer architecture.

Most relevant to us is work on weakly-supervised action localization from video-level labels~\cite{huang2016connectionist, wang2017untrimmednets, nguyen2018weakly}. E-CTC~\cite{huang2016connectionist} proposed an extended CTC framework enforcing frame similarity consistency. Other works~\cite{wang2017untrimmednets, nguyen2018weakly} sample and classify key proposals from untrimmed videos to generate video-level labels. Our work is related to~\cite{huang2016connectionist} in the learning architecture design but in addition to inferring video-level labels, we are able to leverage our learnt concept representation for synthesis and editing. Since most of these methods operate directly on frame-level input which is data-intensive, we use a variant of CTC~\cite{huang2016connectionist} that operates on the pose-level inputs as our baseline for recognition and localization performance.

\xhdr{Human motion and video synthesis.} Methods performing human motion synthesis either use motion graphs~\cite{kovar2008motion, min2012motion}, RNNs~\cite{li2017auto, fragkiadaki2015recurrent, martinez2017human, li2021learn} or autoencoders~\cite{habibie2017recurrent}. Most relevant are works that perform a) action-conditioned motion synthesis and b) controlled motion synthesis from descriptions. 

Action-conditioned motion synthesis is the problem of generating natural  and diverse motion from a given action class. Action2Motion~\cite{guo2020action2motion} proposes a VAE approach and uses Lie algebra to represent natural human motion. ACTOR~\cite{petrovich2021action} generates SMPL human bodies using a Transformer VAE. In contrast, we learn a primitive distribution over curves as our concept representation for each action class. For action-condition motion synthesis, we sample this distribution to generate novel sequences. 

Recent studies~\cite{plappert2016kit, plapper2018synt, lin2018generating, ahuja2019language2pose, ghosh2021synthesis} have looked at controlled motion synthesis from textual descriptions. Most approaches use an RNN-based encoder-decoder model to map descriptions to poses with learning techniques proposed to improve generation quality. In contrast, we perform controlled motion synthesis in a modular way by synthesizing poses for individual concepts and stitching them together, which enables scaling to larger and novel descriptions. Finally, we close the loop by synthesizing realistic videos from the generated motion using prior work in skeleton-to-video synthesis~\cite{chan2019everybody, ren2020human}.

\xhdr{Primitive-based representations for vision.} Several recent works have combined deep neural networks with symbolic structures such as programs to capture higher-level structures. This has been demonstrated on hand-drawn sketches~\cite{ellis2018learning}, natural images~\cite{young2019learning, mao2019program} and videos~\cite{motion2prog}. We build on previous work~\cite{motion2prog} and extend it to capture more expressive motion.

Learning of visual concepts in deep networks has been demonstrated in other domains~\cite{mao2019neuro, tian2019learning, sharma2018csgnet, li2017grass}. NS-CL~\cite{mao2019neuro} maps image patches and words to a joint embedding space for improved performance and data-efficiency in VQA. Similarly, several works in 3D understanding~\cite{tian2019learning, sharma2018csgnet, li2017grass} exploit structure and repetition in 3D shapes to infer concepts such as \textit{table top} and \textit{chair leg}. We extend this idea to human motion, by learning concepts for exercises by exploiting their repetitive nature. %
\vspace{-0.5em}
\section{Motion Concept Learning Framework}
\vspace{-0.5em}
\label{sec:framework}

Our goal is to learn a vocabulary of concept representations that supports video description and synthesis tasks.
We define the description $L$ of a video $V$ as a {\it hierarchical} abstraction at three different levels: primitive level, concept level, and program level. Depicted in \fig{fig:teaser-figure}(I), a {\it motion primitive} is a parameterized trajectory of the subject's joints. A {\it motion concept} is composed of one or more motion primitives, associated with a linguistic name such as {\it jumping jacks} and {\it squats}. A video can be described at the {\it program} level as a sequence of motion concepts, each of which is further concretized as a subsequence of motion primitives.

Our representation not only facilitates human interpretation and analysis of the input video, but also supports interactive  editing tasks. Shown in \fig{fig:teaser-figure}(II), our framework supports interactive editing of the input video at all three levels: users can tweak the parameters for individual primitives, change the concept labels of existing motion concept segments, or even add and remove motion concepts in the video.

Although these tasks have been studied previously,
we focus on creating a unified system for human motion description and synthesis using a small amount of human annotation. For example, the annotation of each concept in our dataset uses weak labels from $<15$ videos on average. 

In the following section, we introduce our framework, {\it \model}, a data-efficient learning framework for human motion description and synthesis.
\modelshort addresses the challenging data-efficient learning problem by leveraging the hierarchical structure of human motions as well as the repetitive structure of concepts in videos. We start from reviewing primitive-based representations for human motion.

\vspace{-0.5em}
\subsection{Primitive-based Motion Representation}
\vspace{-0.5em}
\label{sec:primitives-intro}
Programmatic motion primitives, first presented in~\cite{motion2prog}, are a primitive-based motion representation that capture a level of abstraction higher than pose sequences. The key idea is representing motion trajectories as a sequence of sub-trajectories, each of which can be described with few parameters. Kulal~\etal~\cite{motion2prog} focus on three families of primitives: stationary, linear, and circular. The goal is to fit the motion parameters to well-approximate the input trajectory. The inference of motion primitive sequences involves a dynamic programming algorithm that segments the input pose trajectories at the best possible locations. The recurrence relation for the best fit sequence of primitives for the first $n$ frames is given as $\mathrm{Prims}_{\text{n}} = \min\limits_{k < n} \left[ \mathrm{Prims}_{\text{k}} + \mathrm{synt}(\text{poses}[k:n]) + \lambda \right]$.

Here, $\mathrm{Prims}_{\text{n}}$ is the best fit sequence of primitives for the first $n$ frames and $\mathrm{synt}(\text{poses}[k:n])$ is the best-fit single primitive synthesized for frames $k$ to $n$. The parameter $\lambda$ controls the granularity of the synthesized primitives.

\begin{figure}[t]
  \centering
  \vspace{-.5em}
  \includegraphics[width=0.9\linewidth]{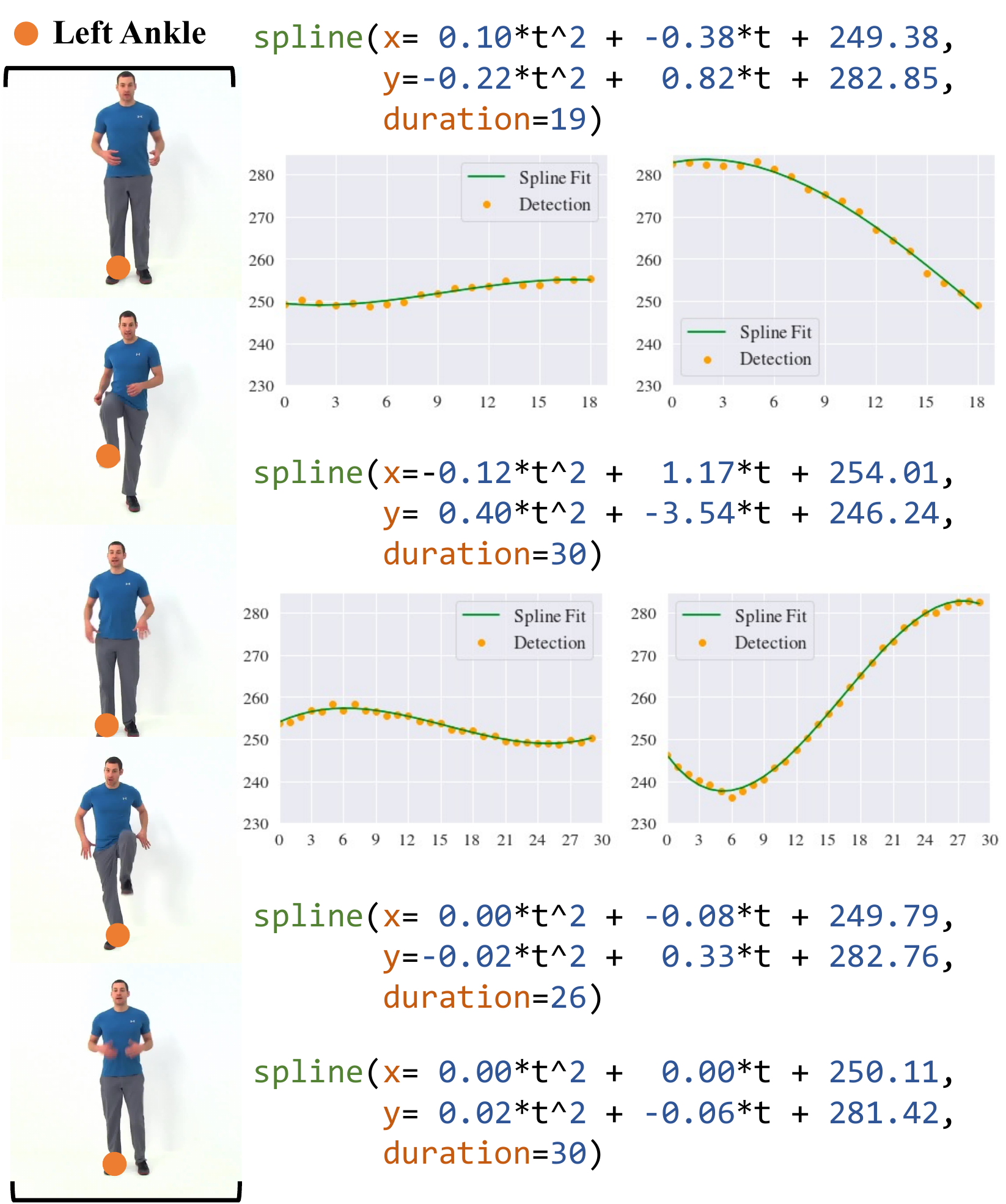}
  \vspace{-0.6em}
  \caption{Four detected motion primitives in a High Knees clip. The third order coefficient is small in all four primitives (absolute value is smaller than $0.01$) so we omitted them in the visualization.}
  \vspace{-1.4em}
  \label{fig:primitive-figure}
\end{figure}

\begin{figure}[t]
  \centering
  \includegraphics[width=\linewidth]{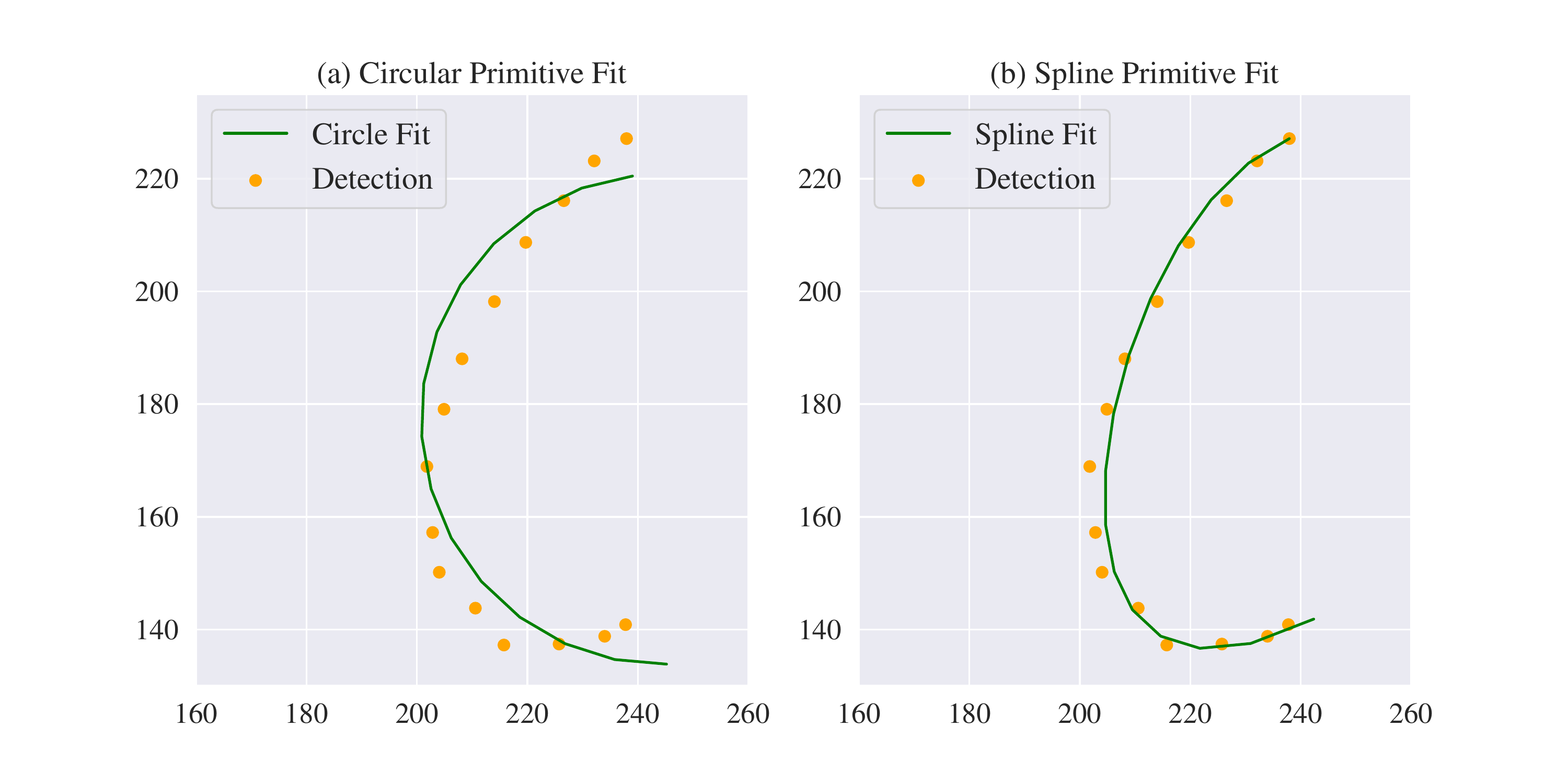}
  \vspace{-2em}
  \caption{Refined spline primitives can better capture the right-wrist motion in the Jumping Jacks class compared to circular primitives.}
  \vspace{-1em}
  \label{fig:spline-figure}
\end{figure}

\vspace{-0.5em}
\subsection{Refined Spline-based Primitives}
\vspace{-0.5em}
\label{sec:primitives-spline}
Motion primitives as defined above do well at representing human motions in a wide variety of videos. However, there are multiple failure modes that make it inapplicable for our use case of generic human exercise videos. Specifically,  some human joint trajectories do not fall into any of the linear/circular/stationary categories. We introduce spline-based primitives which are more general curves than lines and circles while still having only a small number of parameters to learn. We define our spline primitive as
\begin{align}
\mathrm{spline} = \begin{bmatrix}
X = a_{x}t^3 + b_{x}t^2 + c_{x}t + d \\
Y = a_{y}t^3 + b_{y}t^2 + c_{y}t + d \\
T = t
\end{bmatrix}.
\label{eqn:spline-repr}
\end{align}

\fig{fig:primitive-figure} shows the synthesized motion primitives in a high knees exercise. We present more evaluations of these primitives in Section~\ref{sec:exp-spline}. We generalize the motion primitive language of~\cite{motion2prog} to spline-based primitives and use their fast parallel inference algorithm. Figure~\ref{fig:spline-figure} demonstrates a concrete example of the improved expressiveness. The raw input trajectory is a path traced by the right-wrist keypoint in a jumping jacks video. We observe that the spline primitive on the right  approximates this trajectory much better than the best-fit circular primitive. 

\begin{figure*}[t]
  \centering
  \vspace{-0.5em}
  \includegraphics[width=\linewidth]{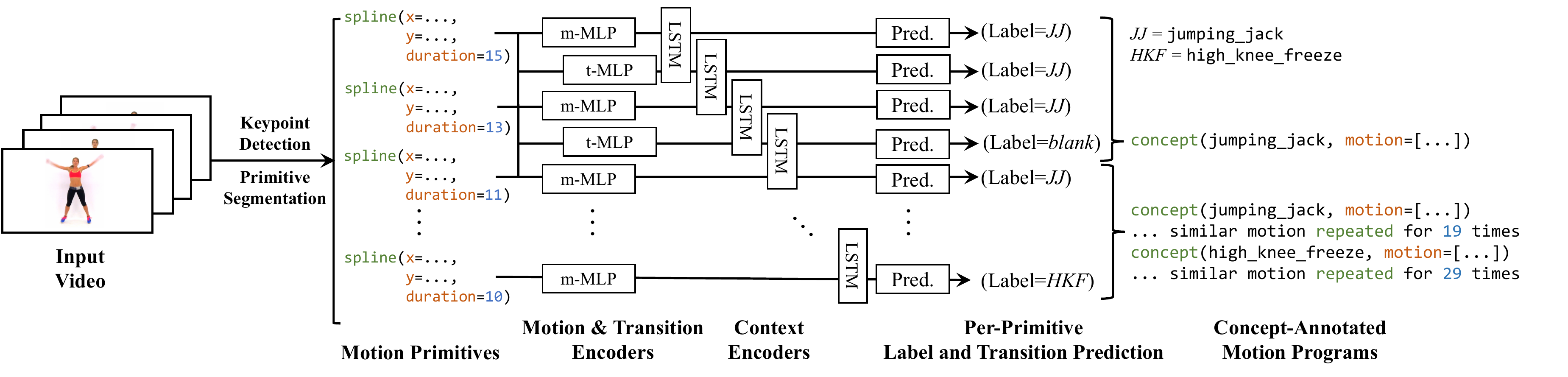}
  \vspace{-1.5em}
  \caption{The model for generating motion descriptions. Given the input video, we first generate motion primitives and feed in the primitives to a simple neural architecture containing feed-forward and recurrent layers to compute concept label predictions for each primitive. We use a connectionist temporal classification-style (CTC) objective to extract the most probable motion description.}
  \vspace{-1.5em}
  \label{fig:arch-figure}
\end{figure*}
 
\subsection{Human Motion Description}
\label{sec:motion-description}
The pipeline in \modelshort for inferring human motion descriptions $L$ from videos $V$ is illustrated in \fig{fig:arch-figure}. Given the input video as a sequence of 2D images, of length $T$, we first run a human pose estimation process~\cite{alphapose} to extract the trajectories of human joints $\bar{P} = \{p_1, p_2, \cdots p_T\}$. We then use the motion primitive segmentation algorithm described in \ref{sec:primitives-intro} to transform the skeleton sequence into primitive sequences $\bar{S} = \{s_1, s_2, \cdots, s_K\}$. The total number of primitives, $K$, is automatically determined by the motion primitive inference algorithm. Our learning task is completely based on this higher-level representation of motion trajectories.   

Next, we encode each motion primitive with a multi-layer perceptron (m-MLP), which is shared across all primitives.
For each pair of consecutive primitives, we encode their features with a separate multi-layer perceptron (t-MLP), which is also shared across all consecutive pairs. The input to t-MLP is the concatenation of two neighboring primitive parameters.
Encoding the transition between two primitives helps us capture the boundaries of motion concepts. We use the same number of output channels for m-MLPs and t-MLPs.

This step gives us in total $2K - 1$ latent features as a sequence: $K$ steps for motion primitives and $K-1$ steps for consecutive primitive pairs.
We further aggregate the contextual information of neighboring latent vectors by running a recurrent cell (we investigated both LSTM~\cite{hochreiter1997long} and GRU~\cite{cho2014learning}) in a local temporal window (the window size is set to 13 in our experiments).
Finally, we decode a concept label distribution for each step $\bar{C} = \{c_1, c_2, \cdots, c_{2K-1}\}$. Each class label is either a category for the motion concept that this primitive belongs to or a ``blank'' token, indicating that it's not part of a motion concept. We refer to $c_1, c_3, \cdots, c_{2K-1}$ as {\it primitive concept} predictions and $c_2, c_4, \cdots c_{2K}$ {\it transition concept} predictions.

Based on the concept label prediction sequence $\bar{C}$, we generate the final prediction of the motion concepts in the video $\bar{L}$ by first iteratively merging consecutive predictions $c_i$ and $c_{i+1}$ with the same label, and then removing all ``blank'' labels. For example, if $\bar{C} = \{\text{JJ},\text{JJ},\text{JJ},\text{Blank},\text{JJ},\text{JJ},\text{JJ}\}$ (JJ is short for ``jumping jack''), the final output with be $\bar{L} = \{\text{JJ}, \text{JJ}\}$. We name this process as $\bar{L} = \textit{compress}(\bar{C})$. We further denote $\textit{uncompress}(\bar{L}) = \{ \bar{C} | \bar{L} = \textit{compress}(\bar{C}) \}$ as the preimage of $\textit{compress}$, \ie, the set of concept label sequences $\bar{C}$ that leads to the same output sequence $\bar{L}$.

Now we can formally define the probability distribution over all possible output sequences $\bar{L}$. We define $p\left( \bar{L} | \bar{S} \right) = \sum_{\bar{C} \in \textit{uncompress}(\bar{L})} p\left(\bar{C}\right) = \sum_{\bar{C} \in \textit{uncompress}(\bar{L})} \prod_{t=1}^{2K-1} p\left(c_t | \bar{S}\right),$
where $p\left(c_t | \bar{S}\right)$ is the concept label sequence predicted by the neural network.

The inference of the labeling sequence $\bar{L}$ can be mathematically formed as finding the $\argmax_{\bar{L}} p\left( \bar{L} | \bar{S} \right)$, which can be solved with a dynamic programming procedure. We discuss this procedure in our supplementary material.

\xhdr{Remark.} Our inference algorithm, as well as one of the loss functions we use  to train the  neural network model, is similar to the connectionist temporal classification (CTC) model. The essential difference between our use case and a standard CTC model is the inclusion of concept label prediction for each pair of consecutive primitives (t-MLPs). This adaptation is important  for our domain: We deal with videos where a single action repeats multiple times, so we need a mechanism to insert correct ``blank'' tokens between two consecutive repetitions of the same motion concept. 

\vspace{-0.5em}
\subsection{Low-Resource Training of Motion Concepts}
\vspace{-0.5em}
\label{sec:low-resource}
Training neural networks for recognizing motion concepts can be data and annotation-intensive, because we need frame-level concept annotations that are laborious to obtain for videos with multiple repetitions. Our goal is to minimize the amount of annotation effort of the user of our system. While we have included a full description and a demo of our annotation tool in the supplementary material, here we briefly introduces the design choices of our labeling tools and how we leverage different types of labels provided by the user to train our motion description model.

Our labeling process leverages the fact that a single motion concept usually repeats  multiple times in an exercise video, \eg, a person does 20 jumping jacks as a group. Our annotation of the motion concepts is {\it hierarchical}: the annotator first specifies the start frame and the end frame of a ``repetition range'' of the motion concept and assigns a label to it (\eg, {\it jumping jacks}). Next, the person specifies the start and end frames of three ``instance ranges'' of the occurrences in this ``repetition range''.

Our labelling pipeline is efficient and quick to annotate. We include a demo in the supplementary material showing how to annotate the start and end boundaries and three repetition intervals for every action class instance, requiring eight mouse-clicks per video.

Second, we aid the labeling process by supplementing diagrams showing the trajectory of human joints. Typically, a motion concept will start and end at a ``local extremum'' of a trajectory. The annotators can click on the local extremum of the trajectory to accurately locate them.

Next, we  leverage the hierarchical annotation of motion concept sequences and our primitive based representation of human motion to train the entire model.
The first objective we used to train the model is a CTC-style loss. Based on the three instance ranges in a repetition range, we compute the average length of the motion concept occurring in the range and estimate the total number $n$ of  occurrences. Then, we compute a pseudo concept label sequence by repeating the concept label $n$ times. We denote the generated label sequence as $\bar{L}^{\dagger}$. Our first loss function is:
$ \mathcal{L}_{\text{CTC}} = -\log p\left(\bar{L} = \bar{L}^{\dagger} | \bar{S}\right).$

We also enforce that, all concept label predictions corresponding to a motion primitive in a repetition range (\ie, the primitive concept predictions, in contrast to the ones associated with each pair of consecutive primitives) should output the concept label associated with the repetition range, $C$. Let $K$ be the number of motion primitives; these predictions are $c_1, c_3, \cdots, c_{2K-1}$. Our second loss function is: $ \mathcal{L}_{\text{P}} = -\log \prod_{i=1,3,\cdots,(2K-1)}  p\left(c_i = C | \bar{S}\right).$

We use a similar idea to generate pseudo-labels for all concept label predictions corresponding to each pair of consecutive primitives (\ie, the transition concept predictions). Within each repetition range, based on the average length of occurrence of motion concept $C$, we segment the motion primitive sequences at the locations that are close to any multiples of the average length. Next, we generate the pseudo-label sequence for transition concept predictions: $c^{\dagger}_2, c^{\dagger}_4, \cdots, c^{\dagger}_{2K}$. $c^{\dagger}_{2i} = C$ if $s_i$ and $s_{i+1}$ belongs to the same segment, and $c^{\dagger}_{2i} = \textit{blank}$ otherwise. The final loss function is:
$\mathcal{L}_{\text{T}} = -\log \prod_{i=2,4,\cdots,2K} p\left(c_i = c^{\dagger}_i | \bar{S}\right).$

The final loss function is $\mathcal{L} = \mathcal{L}_{\text{CTC}} + \lambda_1 \mathcal{L}_{\text{P}} + \lambda_2 \mathcal{L}_{\text{T}}$. Initially, $\lambda_1 = \lambda_2 = 1$. The purposes of $\mathcal{L}_{\text{P}}$ and $\mathcal{L}_{\text{T}}$ are to bias the network towards a preferred solution. Since they are approximate, we set $\lambda_1 = \lambda_2 = 0$ after certain epochs.

\begin{figure}[t]
  \centering
  \includegraphics[width=\linewidth]{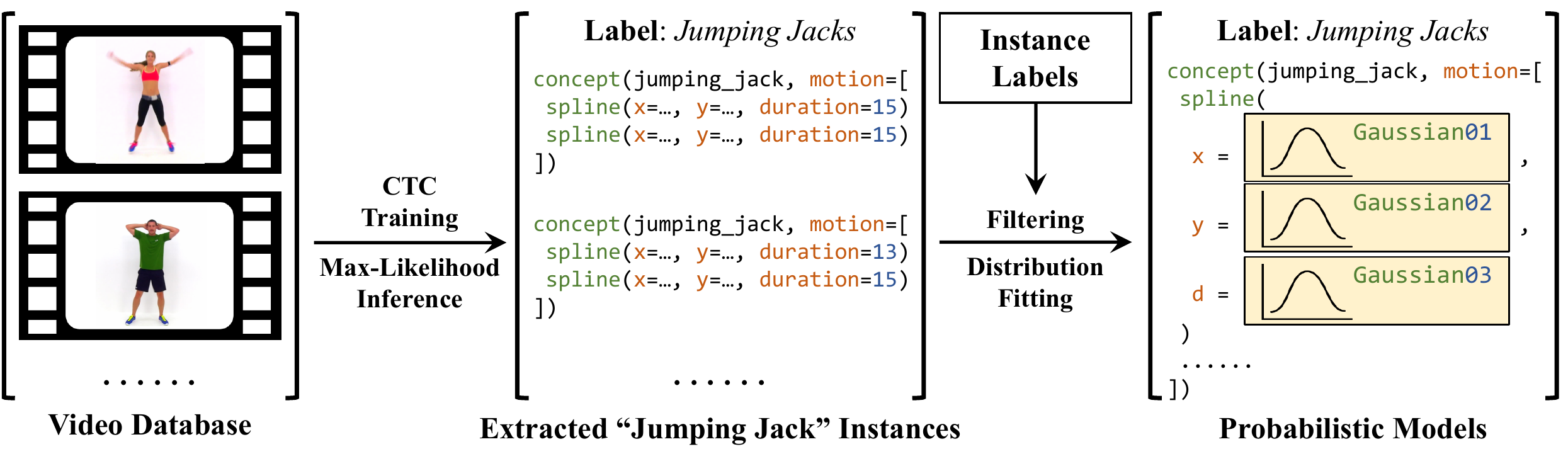}
  \vspace{-1em}
  \caption{Learning a generative model of motion concepts. We leverage the learnt alignments to extract all single repetition examples for each action class. By taking few ground-truth repetitions as reference, we filter them to resolve ambiguities and then learn a primitive distribution as our concept representation.}
  \vspace{-1.5em}
  \label{fig:concept-figure}
\end{figure} 
\vspace{-0.5em}
\subsection{Human Motion Synthesis}
\vspace{-0.5em}
\label{sec:motion-synthesis}
Once we have a learnt model for video description, we can also infer the most probable labels and corresponding alignments for all videos in the dataset, which is done in two steps. Following the notation in \sect{sec:motion-description}, first, given the motion primitive sequence $\bar{S}$, we compute $\bar{L}^* = \argmax_{\bar{L}} p\left( \bar{L} | \bar{S} \right)$, which is the optimal label sequence associated with the input primitive sequence. Next, we compute $\bar{C}^* = \argmax_{\bar{C} \in \textit{uncompress}(\bar{L})^*} p(\bar{C})$, which is the most probable (uncompressed) concept label sequence that yields the label sequence $\bar{L}^*$. This essentially gives us a label associated with each primitive and thus the segmentation of the input primitive sequence $\bar{S}$ into segments, where each segment forms a motion concept.

We use these alignments to extract individual repetitions for all motion concepts from the training dataset. Thus, for each motion concept $c$, we obtain a dataset of occurrences $D_c = \{ d_1, d_2, \cdots \}$ where each element $d_i$ corresponds to a small segment composed of a sequence of spline curves $d_i = \{ s_i^1, s_i^2, \cdots s_i^{l_i} \}$. We then perform two steps of filtering of this dataset $D_c$. First, we compute mode of the number of splines used in instances of this concept $l_c^* = \textit{mode}\left(\{l_i\}\right)$ and then filter out all $d_i$'s whose number of splines is not $l_c^*$. Furthermore, we threshold and choose primitives that agree well with the single repetition examples already annotated for concept $c$. Denote the filtered set of occurrences for concept $c$ as $D'_c$. Each element in $D'_c$ has the same number of spline curves. We learn a distribution for each spline curve by fitting a simple Gaussian model over the parameters. We can then sample these distributions to generate novel instances of concept $c$. This pipeline is visualized in Figure~\ref{fig:concept-figure}. The details of the model fitting and data filtering are included in the supplementary material.

Based on the learned generative model for motion concepts, we can perform both action-conditioned motion synthesis and controlled motion synthesis from descriptions. To perform action-conditioned motion synthesis, we sample the learnt primitive distributions for the corresponding action.

To perform controlled motion synthesis given our descriptions, for example \textit{four jumping jacks followed by three squats}, we sample the learnt distributions for the required number of distributions in the given order of actions. We stitch these segments together by manually setting the start pose of a segment to the last pose of the previous segment for smooth transitions. Hence, our approach is modular and scales better to longer descriptions than previous methods.

\vspace{-0.5em}
\section{Experiments}
\vspace{-0.5em}
\label{sec:experiments}
We conduct multiple experiments to evaluate the performance of our framework. First, we compare the spline-based primitives with the primitives proposed in previous work~\cite{motion2prog} in Section~\ref{sec:exp-spline}. We  evaluate \modelshort on three downstream applications: i) human motion description in Section~\ref{sec:exp-detection}, ii) action-conditioned motion synthesis in Section~\ref{sec:exp-single-action-synthesis}, and iii) controlled motion and video synthesis from descriptions in Section~\ref{sec:exp-synthesis}. Qualitative results on interactive editing can be found in the supplementary material. All downstream applications are evaluated on the MotiCon dataset. 

\subsection{MotiCon Dataset}
\label{sec:dataset}

We have gathered a dataset of human cardio videos to evaluate the concept learning framework and the downstream tasks. We scraped YouTube for single human exercise videos and filtered these videos to those with a stable single camera, 1920$\times$1080 resolution and 30 FPS. We label 6 action classes with the interval and single rep annotations as mentioned earlier.  The statistics of the final dataset are in Table~\ref{tab:data-statistics}. We refer to this dataset as MotiCon in our experiments.\footnote{Our dataset contains videos with Standard YouTube license. It is available on our website as a list of URLs with timestamps and annotations.}

\begin{table}[t]
\caption{MotiCon dataset statistics. For each class, we list \textbf{\# vids} (total number of videos), \textbf{\# mins} (total number of minutes) and \textbf{\# reps} (total number of repetitions).}
\small
\vspace{-1em}
\setlength{\tabcolsep}{10pt}
    \centering
    \begin{tabular}{lccc}
        \toprule
        Class  & \# vids & \# mins & \# reps \\
        \midrule
        Arm Cross Swings &	12 & 6.68 &	127 \\
        Buttkickers & 9 &	5.94 &	425 \\
        High Knee March & 8 &	3.70 &	81 \\
        Jumping Jack & 23 &	13.03 &	853 \\
        Toe Touch Sweeps & 11 &	5.00 &	35 \\
        Torso Twists & 8 &	3.90 &	79 \\
        \bottomrule
    \end{tabular}
    \label{tab:data-statistics}
    \vspace{-6pt}
\end{table}

\subsection{Evaluation of Spline Primitives}
\label{sec:exp-spline}

We use two  metrics. First, we are interested in how well the primitive-based representation encodes the ground-truth poses. We execute our primitive-based representation to generate poses at each time step and compute the keypoint difference (KD) of the generated and ground-truth poses. We report KD as L2 error per joint per frame averaged across all videos (lower is better), and the average time for primitive synthesis in seconds (Time, lower is better).

\begin{table}[t]
\vspace{-0.5em}
\caption{Evaluating expressive power of motion primitives. On two datasets, we compute \textbf{KD} (keypoint difference w/ ground-truth), and \textbf{Time} (synthesis time in seconds).}
\small
\vspace{-1em}
\setlength{\tabcolsep}{5pt}
    \centering
    \begin{tabular}{lcccc}
        \toprule
        \multirow{2}{*}{Method} & \multicolumn{2}{c}{GolfDB\cite{golfdb}} & \multicolumn{2}{c}{MotiCon} \\
        \cmidrule(lr){2-3}\cmidrule(lr){4-5} 
         & KD$\downarrow$ & Time$\downarrow$ & KD$\downarrow$ & Time$\downarrow$ \\
        \midrule
        Motion Programs\cite{motion2prog} & 1.171\% & 48.82 & 0.575\% & 57.65 \\
        PMC (Ours) & \textbf{0.529\%} & \textbf{3.22} & \textbf{0.229\%} & \textbf{15.92}  \\
        \midrule
        Ground-Truth & 0\% & \ding{55} & 0\% & \ding{55} \\ 
        \bottomrule
    \end{tabular}
    \label{tab:spline-statistics}
    \vspace{-12pt}
\end{table}

\xhdr{Analysis.} We evaluate on two datasets: GolfDB~\cite{golfdb}\footnote{GolfDB is released with a CC BY-NC 4.0 license.} and MotiCon.  The results are in Table~\ref{tab:spline-statistics}. We observe that our spline-based representation decreases KD by $\sim$ 57\%. We also see significant qualitative improvement in the results as our representation can faithfully encode a large range of motions. We supply these results in our supplementary material. Our synthesis times are also lower than Motion Programs because synthesizing best-fit circular primitives is more expensive than synthesizing splines. However, spline-based representations can require more parameters to represent the primitives, at times  1.5-2x of Motion Programs.

\subsection{Human Motion Description}
\label{sec:exp-detection}

We evaluate the task of human motion description, which is to infer the correct order of concept sequences along with their localization in unseen videos.

 We use two metrics. First, we compute the average NormED (normalized edit distance), which is the edit distance of the inferred concept sequence with the ground-truth concept sequence normalized by the maximum length of the two sequences (lower the better). We define SeqAcc (sequence accuracy) as $(1 - \textnormal{NormED}) \times 100\%$ to translate the NormED to an accuracy metric. Finally, to evaluate the localization accuracy, we compute the repetition mAP (mAP). We annotate the ground-truth action localization for the test set and evaluate the average mAP of the inferred localization for varying IoU thresholds (mAP@[.5:.95]~\cite{caba2015activitynet}, higher is better). SeqAcc reflects recognition performance while mAP reflects the localization performance.

\xhdr{Baselines.} We compare multiple architectures with our approach. First, we compare to vanilla encoder-decoder models (prefixed seq2seq) to show that CTC-based approaches are superior for this task. For CTC-based architectures, we compare the effects of using different kinds of recurrent layers: GRU and LSTM. For all models, we implement two variants, one taking as input the pose sequences (suffixed -Pose) and second taking as input the primitive sequences (suffixed -Prim). We also study the effects of various loss terms by ablating the primitive loss $\mathcal{L}_P$ and transition loss $\mathcal{L}_T$.

\begin{table}[tp]
\caption{Evaluating human motion description. We compute the \textbf{NormED} (normalized edit distance) and \textbf{SeqAcc} (sequence accuracy) for recognition  and \textbf{mAP} (repetition mAP) for localization.}
\vspace{-1em}
\small
\setlength{\tabcolsep}{5pt}
    \centering
    \begin{tabular}{lccc}
        \toprule
        Method & NormED$\downarrow$ & SeqAcc$\uparrow$ & mAP$\uparrow$ \\
        \midrule
        seq2seq-LSTM-Pose & 0.64840 & 35.16\% & \ding{55}  \\
        seq2seq-GRU-Pose & 0.31353 & 68.65\% & \ding{55}  \\
        seq2seq-LSTM-Prim & 0.60810 & 39.19\% & \ding{55}  \\
        seq2seq-GRU-Prim & 0.59770 & 40.23\% & \ding{55}  \\
        \midrule
        LSTM-Pose & 0.22990 & 77.01\% &  0.00\% \\
        GRU-Pose & 0.18057 & 81.94\% & 0.00\%  \\
        \midrule
        GRU-Prim w/o $\mathcal{L}_P$ \& $\mathcal{L}_T$ & 0.11072 & 88.93\% & 0.10\%  \\
        GRU-Prim w/o $\mathcal{L}_P$ & 0.10303 & 89.70\% & 20.58\%   \\
        GRU-Prim w/o $\mathcal{L}_T$ & 0.14763 & 85.24\% & 17.87\%  \\
        \midrule
        LSTM-Prim (Ours) & 0.08831 & 91.17\%  & \textbf{45.43\%}  \\
        GRU-Prim (Ours) & \textbf{0.08469} & \textbf{91.53\%}  & 38.74\%  \\
        \bottomrule
    \end{tabular}
    \label{tab:human-motion-description}
\end{table}

\begin{figure}[tp]
  \centering
  \vspace{-1em}
    \includegraphics[width=\linewidth]{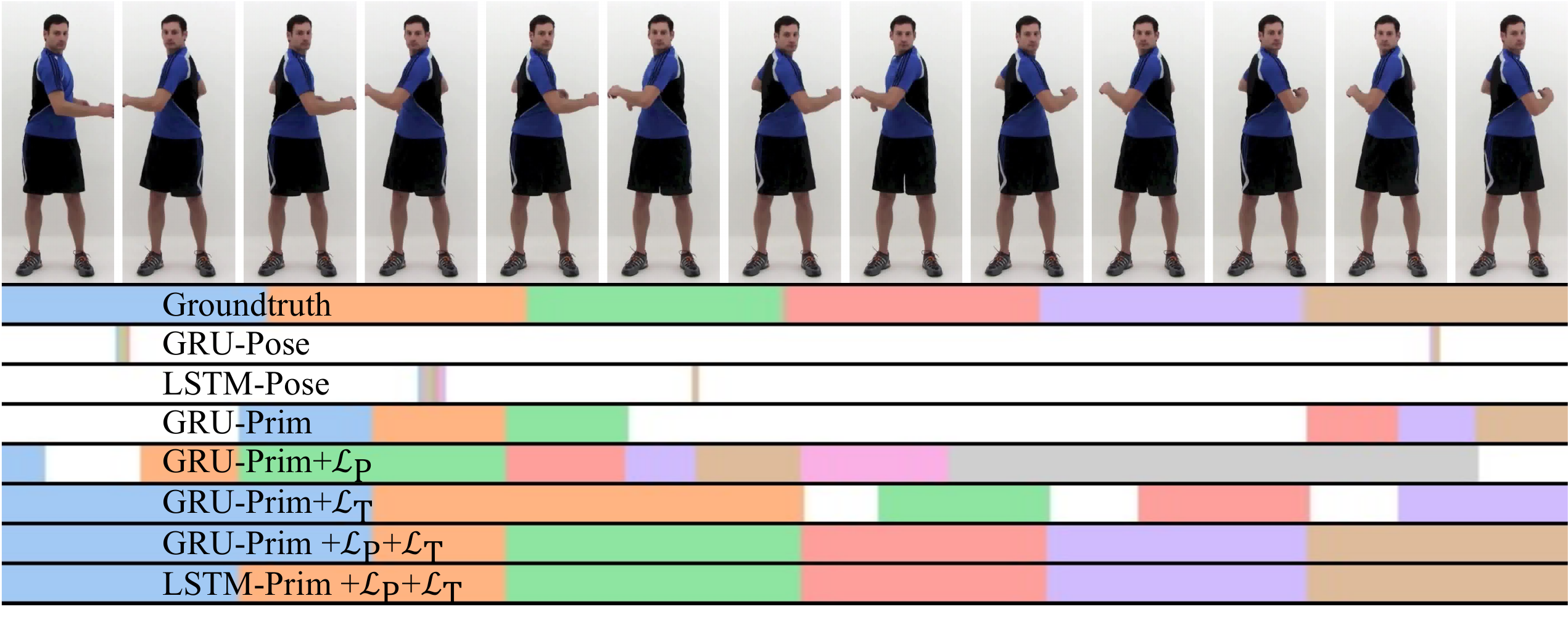}
  \vspace{-1.5em}
    \caption{Visualising the localized ``torso twists'' concept in a video. Different colors represent different segments of the localized by models. Even if they all predict the correct number of repetitions of ``torso twists'', models using our primitive-based representations and $\mathcal{L}_\text{P}+\mathcal{L}_\text{T}$ finds the segments that maximally align with the ground-truth.}
  \vspace{-1.5em}
    \label{fig:short-a}
\end{figure}

\xhdr{Analysis.}  The results of the evaluation are in Table~\ref{tab:human-motion-description}. We present the average over three independent runs for each model. We observe that CTC models perform better than seq2seq models for recognition. CTC models are also able to perform localization by design while seq2seq models cannot but we see that performing localization from pose level inputs is harder than primitive level. We also observe that our models outperform all baselines on both the recognition and localization metrics. A visualization of the localized intervals for all models is presented in Figure~\ref{fig:short-a}, where different colors denote different intervals located by the models. The localized intervals from our model aligns best with the ground truth, which yields more accurately localized motion descriptions and also facilitates the learning of the motion generative model. In both Table~\ref{tab:human-motion-description} and Figure~\ref{fig:short-a}, we study effects of ablating our loss terms. We see that both losses contribute to improved performance, performing the best when used in combination.

\vspace{-0.5em}
\subsection{Action-Conditioned Motion Synthesis}
\vspace{-0.5em}
\label{sec:exp-single-action-synthesis}

We evaluate our concept representation on action-conditioned motion synthesis. We use the localized segments from our learnt description model to learn our concept representation. We sample this representation to generate novel instances from individual action classes.

As in Guo~\etal~\cite{guo2020action2motion}, we evaluate on four quantitative metrics that together try to capture the correctness and diversity of the synthesized motions. We compute the i) FID score between the generated motions and real motions, ii) recognition accuracy (Acc) of the generated motions, iii) diversity (Div) defined as the variance of generated motion across action classes (closer to real motions is better), and iv) multimodality (MM) defined as the variance of generated motions within each action class (closer to real motions is better). We use a RNN-based action classifier as our feature extractor to compute these metrics. More details on the metrics can be found in the supplementary materials.

\xhdr{Baselines.} We compare our method with Act2Mot-2D which is a 2D variant of Action2Motion~\cite{guo2020action2motion} without their Lie algebra variant (as it is 3D-specific). Both the baselines and our method take the same localized segments as input data. For the baseline we implement two variants, one predicting the pose sequences and second predicting the primitive sequences (suffixed -Prim).

\xhdr{Analysis.} The results of the evaluation are in Table~\ref{tab:single-class-synthesis}. We observe that our method has lower FID score and higher Acc value conveying that the synthesized motions closely resemble the real motion and can be well-classified to the correct classes. Our method also produces diverse motion across classes (Div) but produces less diversity within an action class (MM) compared to the baseline, due to our hyperparameter choice that controls variance in model fitting. One can trade-off a lower value of Acc for improved variance. More details are provided in the model fitting section of the supplementary material. 

\begin{table}[t]
\caption{Evaluating action-conditioned motion synthesis. We compare the \textbf{FID} score, \textbf{Acc} (recognition accuracy), \textbf{Div} (variance across action classes) and \textbf{MM} (variance within action classes).  }
\small
\vspace{-1em}
\setlength{\tabcolsep}{5pt}
    \centering
    \begin{tabular}{lcccc}
        \toprule
        Method & FID$\downarrow$ & Acc$\uparrow$ & Div$\rightarrow$ & MM$\rightarrow$ \\
        \midrule
        Real motions & 0.220 & 98.72\% & 8.363 & 1.891 \\
        \midrule
        Act2Mot-2D~\cite{guo2020action2motion} & 23.465 & 84.39\% & 7.748 & \textbf{2.001} \\
        Act2Mot-2D-Prim & 16.971 & 83.49\% & 7.872 & 2.794 \\
        PMC (Ours) & \textbf{2.406} & \textbf{97.48\%} & \textbf{8.182} & 1.141   \\
        \bottomrule
    \end{tabular}
    \label{tab:single-class-synthesis}
\end{table}

\vspace{-0.5em}
\subsection{Controlled Motion and Video Synthesis}
\vspace{-0.5em}
\label{sec:exp-synthesis}

\begin{table}[t]
\vspace{-0.5em}
\caption{Evaluating controlled motion synthesis. We compare the \textbf{APE} (average position error) and \textbf{AVE} (average variance error).  }
\small
\vspace{-1em}
\setlength{\tabcolsep}{18pt}
    \centering
    \begin{tabular}{lcc}
        \toprule
        Method & APE$\downarrow$ & AVE$\downarrow$ \\
        \midrule
        Lin~\etal~\cite{lin2018generating}   & 0.2470 & 0.0457 \\
        JL2P~\cite{ahuja2019language2pose} & 0.2504 &  0.0440  \\
        Lin~\etal-Prim & 0.2075 & 0.0359 \\
        JL2P-Prim & 0.2052 & 0.0358  \\
        PMC (Ours) & \textbf{0.1531} & \textbf{0.0174}    \\
        \bottomrule
    \end{tabular}
    \label{tab:controlled-motion-synthesis}
    \vspace{-1em}
\end{table}

We evaluate controlled motion and video synthesis given descriptions. For the test set, we take in video-level labels and synthesize the pose and video sequence.

As in prior works~\cite{ahuja2019language2pose, ghosh2021synthesis}, we evaluate the motion synthesis on two key metrics: i) Average Position Error (APE), which measures the positional difference between the generated pose sequence with the ground-truth pose sequence (lower the better), and ii) Average Variance Error (AVE), which measures the difference in variance of the generation with the ground-truth (lower the better).

We train two skeleton-to-human GANs~\cite{chan2019everybody} to synthesize realistic videos from the generated pose sequence for all models. We compute the PNSR, SSIM, LPIPS and MSE metrics to judge the resulting video quality. 

\xhdr{Baselines.} We compare against two methods: i) Lin~\etal~\cite{lin2018learning}, which first trains an autoencoder to learn a human motion model followed by an encoder-decoder model to translate description to pose sequences, ii) JL2P~\cite{ahuja2019language2pose}, which trains an encoder-decoder model with a joint embedding space for descriptions and poses. Additionally, we also modify both methods to predict primitive sequences (denoted with -Prim) for a stronger baseline. 

\xhdr{Analysis.} Motion synthesis evaluation results are in Table~\ref{tab:controlled-motion-synthesis}. We observe that our model outperforms the baselines on both APE and AVE by a fair margin, showing that the generated motion sequence matches the prompted description more faithfully. We also note that augmenting the baselines with our spline primitives improves performance. The results of video synthesis evaluations are in Table~\ref{tab:controlled-video-synthesis}. Similar to motion synthesis, we observe that video synthesized from our method outperforms the baselines on all metrics. Qualitative visualizations are provided in the supplemental material.

\begin{table}[t]
\caption{Evaluating controlled video synthesis. We compare the \textbf{PSNR}, \textbf{SSIM}, \textbf{LPIPS} and \textbf{MSE} metrics for the generated videos. }
\small
\vspace{-1em}
\setlength{\tabcolsep}{6pt}
    \centering
    \begin{tabular}{lcccc}
        \toprule
        Method & PSNR$\uparrow$ & SSIM$\uparrow$ & LPIPS$\downarrow$ & MSE$\downarrow$\\
        \midrule
Lin~\etal~\cite{lin2018generating} & 17.791 & 0.925 & 0.112 & 1230.732 \\ 
JL2P~\cite{ahuja2019language2pose} & 17.898 & 0.926 & 0.109 & 1192.657 \\ 
Lin~\etal-Prim & 18.574 & 0.930 & 0.094 & 1067.636 \\ 
JL2P-Prim & 18.637 & 0.931 & 0.093 & 1067.194 \\ 
PMC (Ours) & \textbf{19.355} & \textbf{0.934} & \textbf{0.085} & \textbf{916.898} \\ 

        \bottomrule
    \end{tabular}
    \label{tab:controlled-video-synthesis}
    \vspace{-1.5em}
\end{table}

\vspace{-0.5em}
\section{Conclusion}
\vspace{-0.5em}
\label{sec:limitations}
We present a single framework for human motion description and synthesis via programmatic motion concepts. We show that our concept representation can be learnt in a data-efficient manner and is effective on multiple applications. We present a brief discussion on limitations and societal impacts of our work in the supplementary material.

\vspace{-1em}

{\small \paragraph{Acknowledgements.} This work is in part supported by a Magic Grant from the Brown Institute for Media Innovation, the Toyota Research Institute, Stanford HAI, Samsung, IBM, Salesforce, Amazon, and the Stanford Aging and Ethnogeriatrics (SAGE) Research Center under NIH/NIA grant P30 AG059307. The SAGE Center is part of the Resource Centers for Minority Aging Research (RCMAR) Program led by the National Institute on Aging (NIA) at the National Institutes of Health (NIH). Its contents are solely the responsibility of the authors and does not necessarily represent the official views of the NIA or the NIH.} 
{\small
\bibliography{motionprog,bpi}
\bibliographystyle{ieee_fullname}
}

\end{document}


\title{Programmatic Concept Learning for Human Motion Description and Synthesis \\ Supplementary Materials}

\author{Sumith Kulal\thanks{\xspace and $^\dagger$ indicate equal contribution. Project page: \url{https://sumith1896.github.io/motion-concepts}}\\
Stanford University\\
\and
Jiayuan Mao\footnotemark[1]\\
MIT \\
\and
Alex Aiken$^\dagger$\\
Stanford University\\
\and
Jiajun Wu$^\dagger$\\
Stanford University\\
}
\maketitle

\vspace{-0.5em}
\section{Details of Data Filtering and Model Fitting}
\vspace{-0.5em}
\label{sec:model-fit}

As mentioned in Section 3.5 of the main paper,  we use the alignments obtained from the description model to extract individual repetitions for all motion concepts from the training dataset. For each motion concept $c$, we obtain a dataset of occurrences $D_c = \{ d_1, d_2, \cdots \}$ where each element $d_i$ corresponds to a small segment composed of a sequence of spline curves $d_i = \{ s_i^1, s_i^2, \cdots s_i^{l_i} \}$. Since this dataset $D_c$ has been obtained from description model predictions, we perform two steps of filtering to generate cleaner data for learning our synthesis model. 

\paragraph{Length Filtering.} First, for each concept $c$, we compute the mode of the number of splines used in instances of this concept, denoted as $l_c^* = \textit{mode}\left(\{l_i\}\right)$ and then filter out all $d_i$'s whose number of splines is not $l_c^*$. It is possible to fit a synthesis model for each length and sample from each of these models but for simplicity we only consider models with a fixed number of primitives per class.

\paragraph{Similarity Filtering.} We use the already annotated single repetition examples as the ground truth reference $\{ g_1, g_2, \cdots \}$. 
We define the distance ($\textit{distance}$) between $d_i$ and $g_j$ as the average $L_2$ distance between four points sampled at equal distance across all spline curves $\{ s_i^1, s_i^2, \cdots s_i^{l_i} \}$ (note that the number of splines in $d_i$ and $g_j$ must match due to the first-step length filtering).
Next, we filter out $d_i$ if $\min_{j}\textit{distance}(d_i, g_j) > F$. Here, $F$ is a hyperparameter which controls the error threshold. We choose $F = 8$ in our experiments.%

\paragraph{Model Fitting.} After these two steps of filtering, we now have dataset $D'_c = \{ d'_1, d'_2, \cdots \}$ where each element $d'_i$ is a sequence of spline curves $d'_i = \{ t_i^{1}, t_i^{2}, \cdots t_i^{l^*_c} \}$.
For each index $\ell$ with $1 \leq \ell \leq l^*_c$, we fit a simple Gaussian model over all occurrences $$G^\ell_c \sim \mathcal{N}(\mu(t^{l}),\, \textit{cov\_f} \cdot \sigma(t^{l})^{2})$$
where $\textit{cov\_f}$ is a hyperparameter that controls the variance of the generated motion. We set $\textit{cov\_f}$ to 0.01 in our experiments and present ablation studies in the following section. To sample new motion instance $\bar{d}$, we sample each of $G_c^l$ in sequence from $l$ set to $1$ to $l^*_c$.

\section{Inference with Dynamic Programming}
\label{sec:dp-infer}
Given a primitive sequence $\bar{S}$, our goal in human motion description is to infer a label sequence $\bar{L}$. As described in Section 3.3 of the main paper, the inference of $\bar{L}$ is equivalent to finding the argmax of $p\left( \bar{L} | \bar{S} \right) = \sum_{\bar{C} \in \textit{uncompress}(\bar{L})} \prod_{t=1}^{2K-1} p\left(c_t | \bar{S}\right),$
where $p\left(c_t | \bar{S}\right)$ is the concept label sequence predicted by the neural network. 

Since there are several possible alignments $\bar{C} \in \textit{uncompress}(\bar{L})$ for a given label sequence $\bar{L}$, the argmax could be quite expensive to compute in a brute-force manner. Hence, we use an efficient dynamic programming approach. The key idea is to compute prefix label sequences for prefix primitive sequences of $\bar{L}$ and merge different alignments that give the same output, which makes the inference and loss computation tractable. In practice, implementations of CTC manage this under-the-hood\footnote{\label{ctc}https://distill.pub/2017/ctc/}.

\section{Ablation of Covariance Factor}
\label{sec:covf-ablate}
\begin{figure}[t]
    \centering
    \includegraphics[width=.495\linewidth]{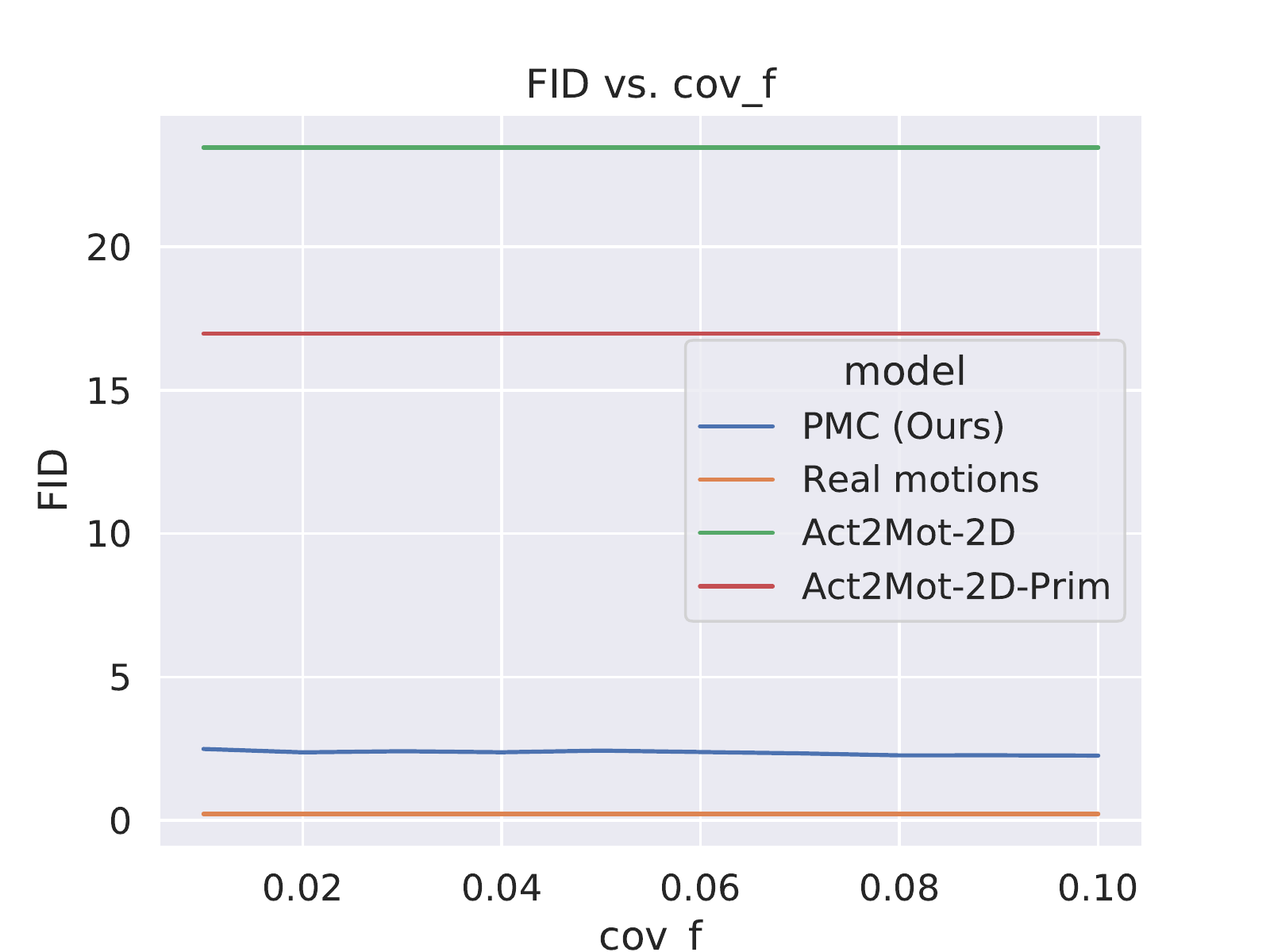}
    \includegraphics[width=.495\linewidth]{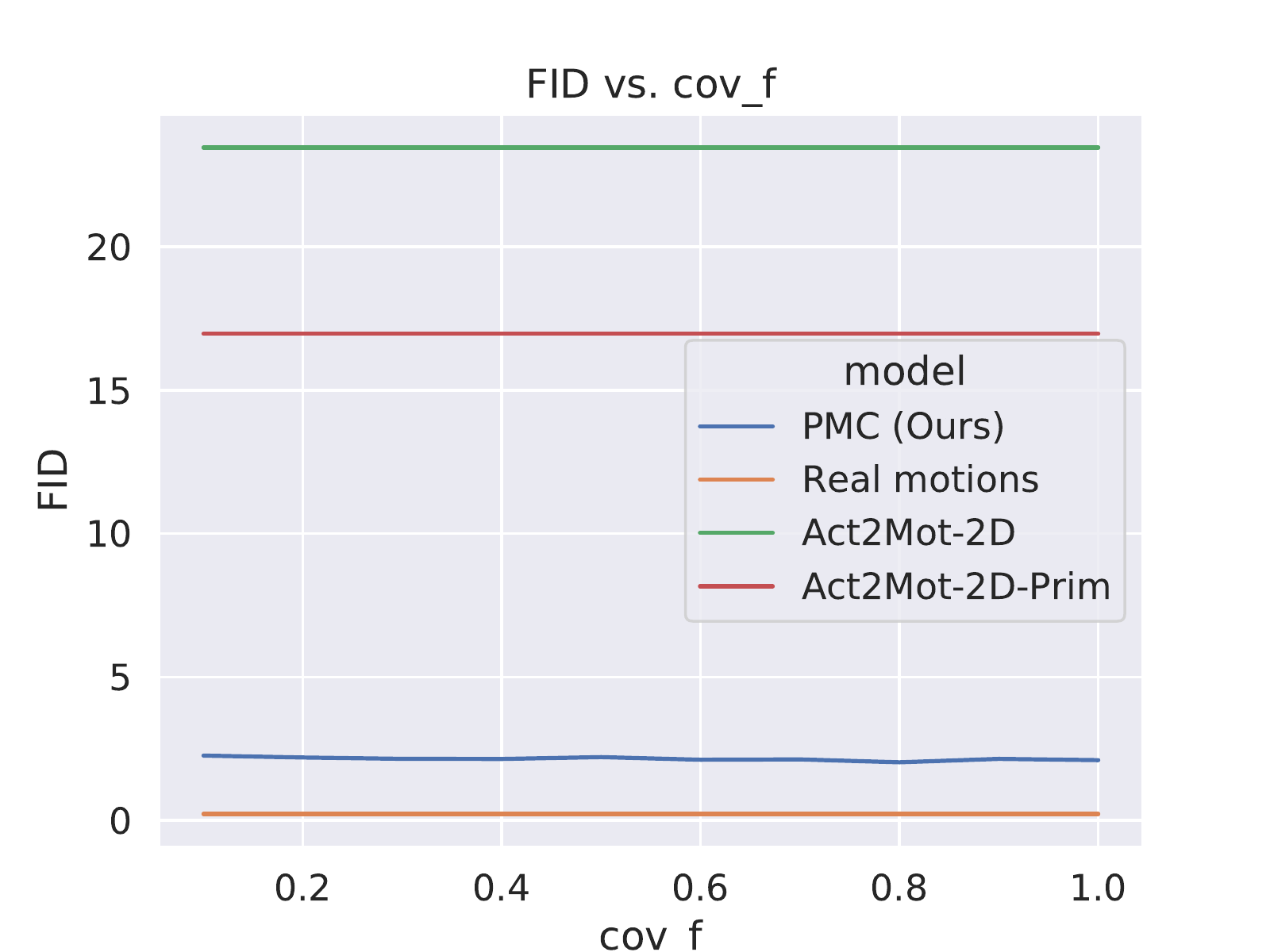}
    \\%
    \vspace{-5pt}
    \includegraphics[width=.495\linewidth]{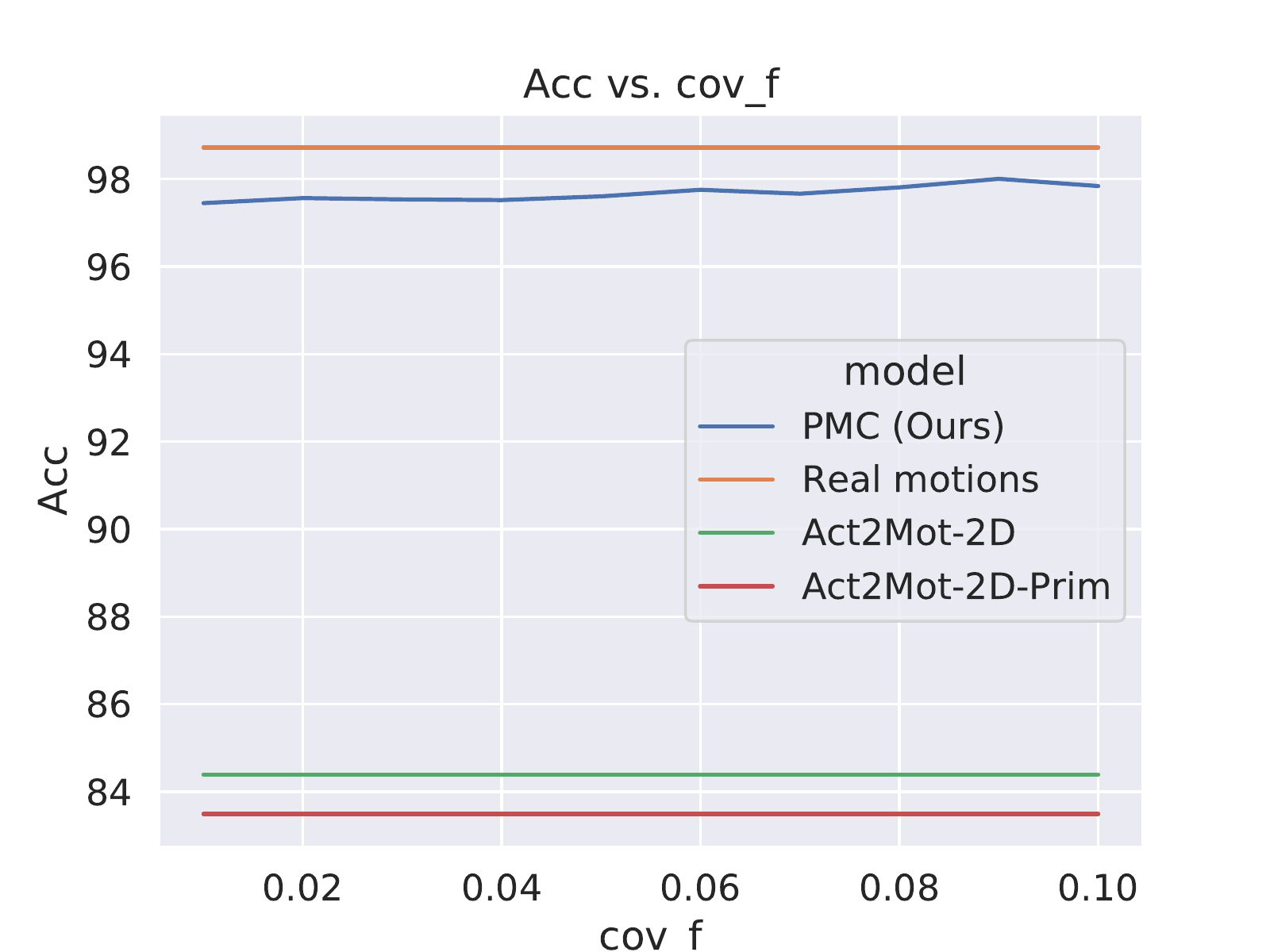}
    \includegraphics[width=.495\linewidth]{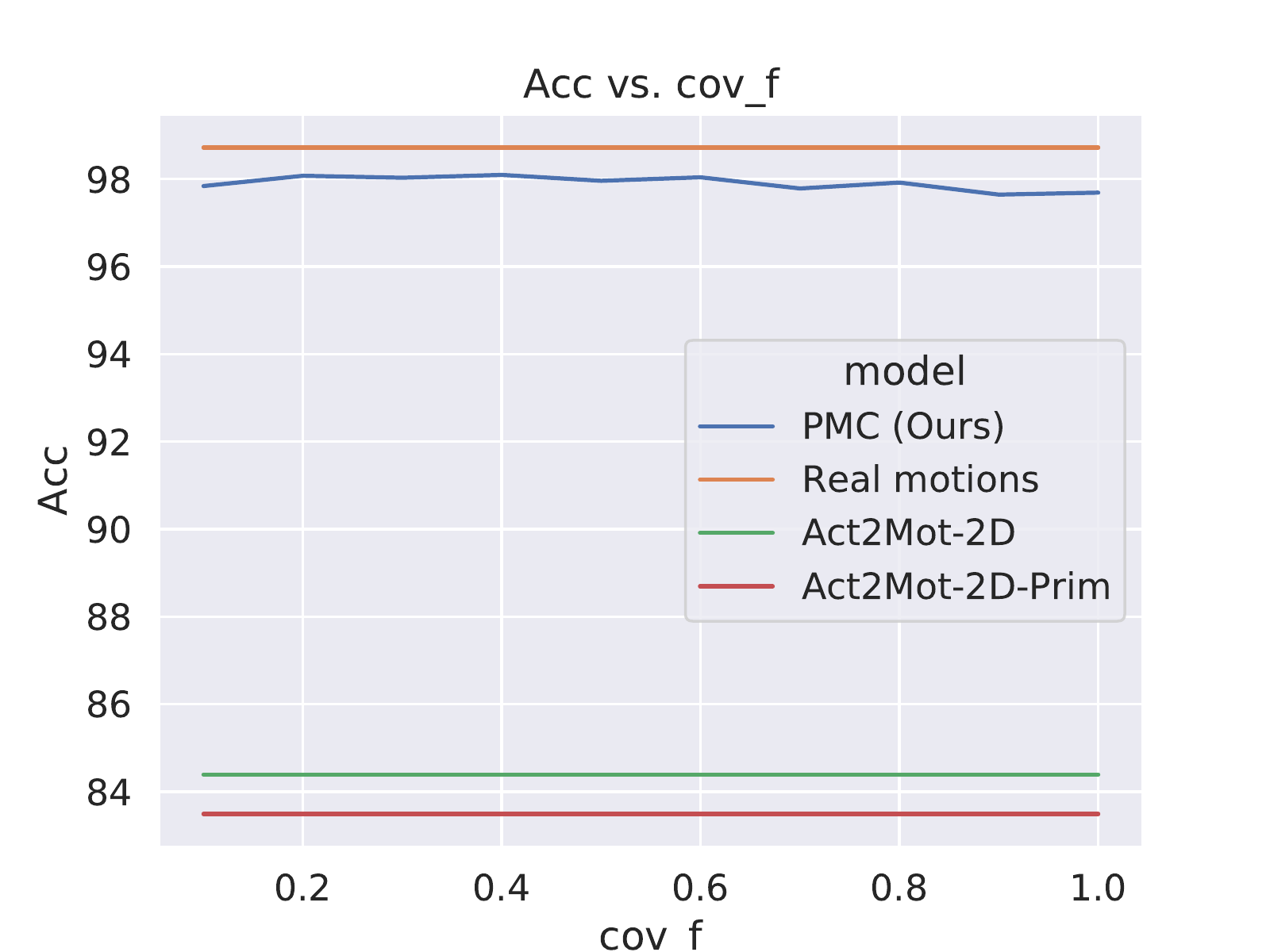}
    \\%
    \vspace{-5pt}
    \includegraphics[width=.495\linewidth]{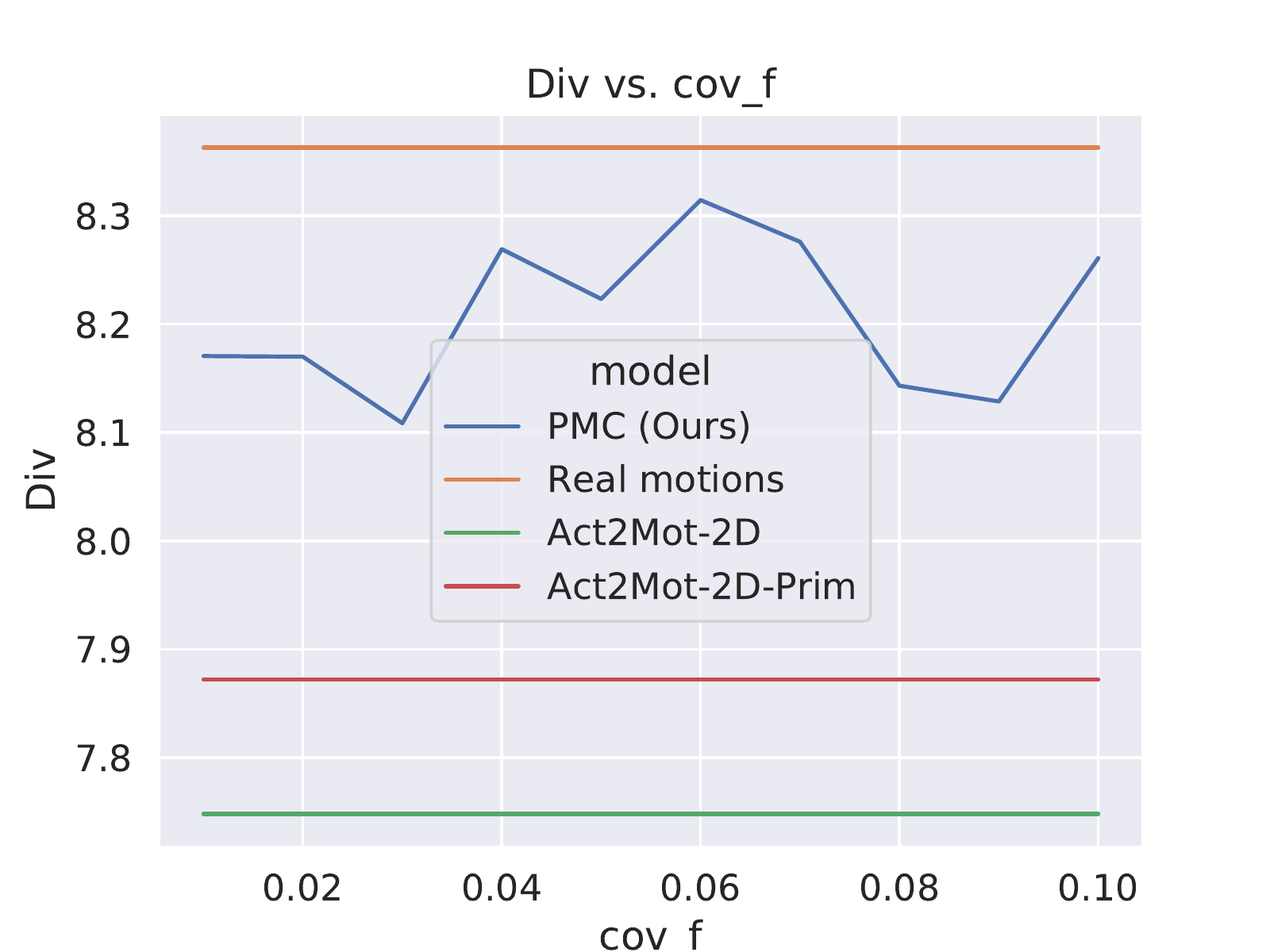}
    \includegraphics[width=.495\linewidth]{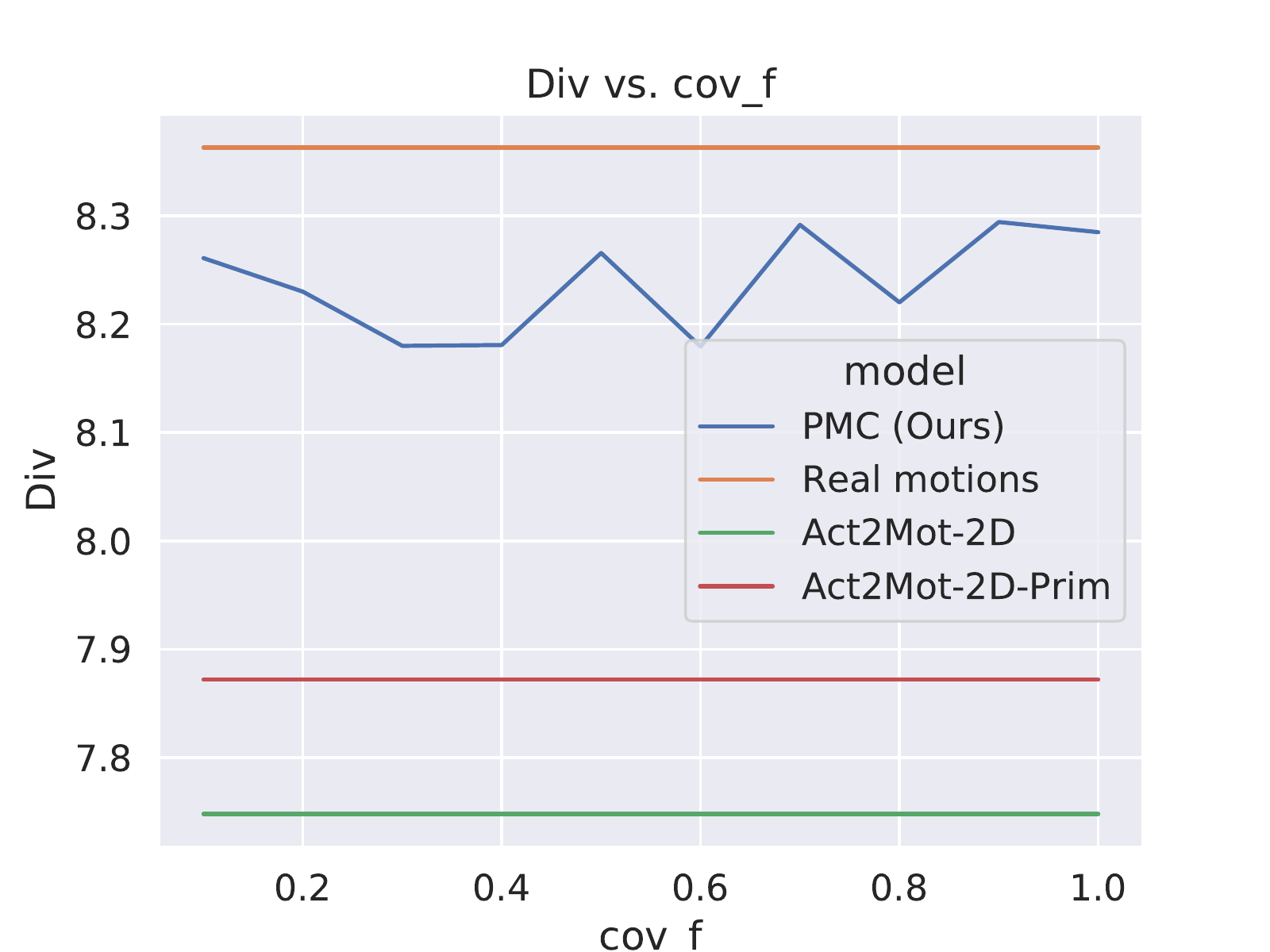}
    \\%
    \vspace{-5pt}
    \includegraphics[width=.49\linewidth]{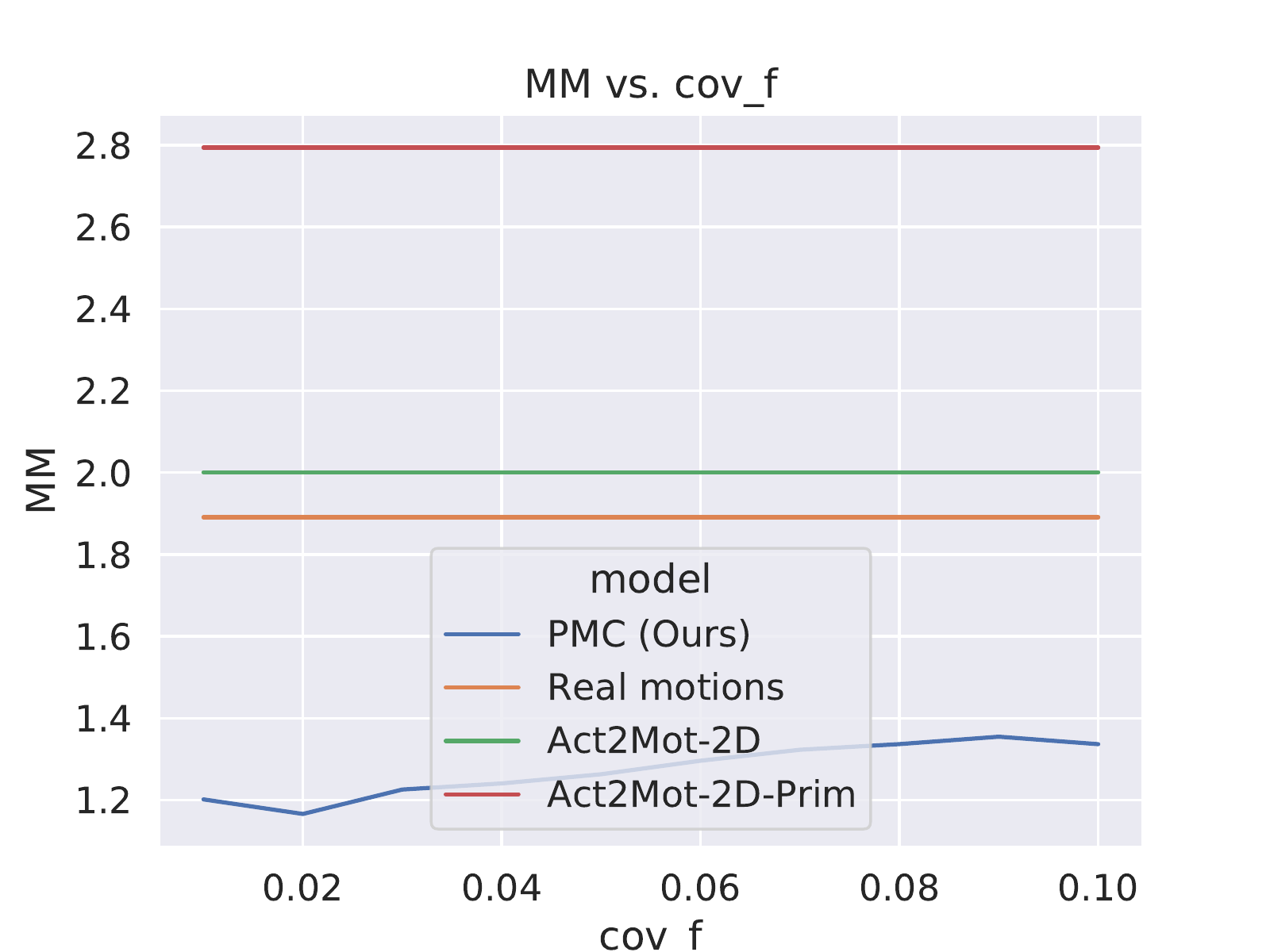}
    \includegraphics[width=.49\linewidth]{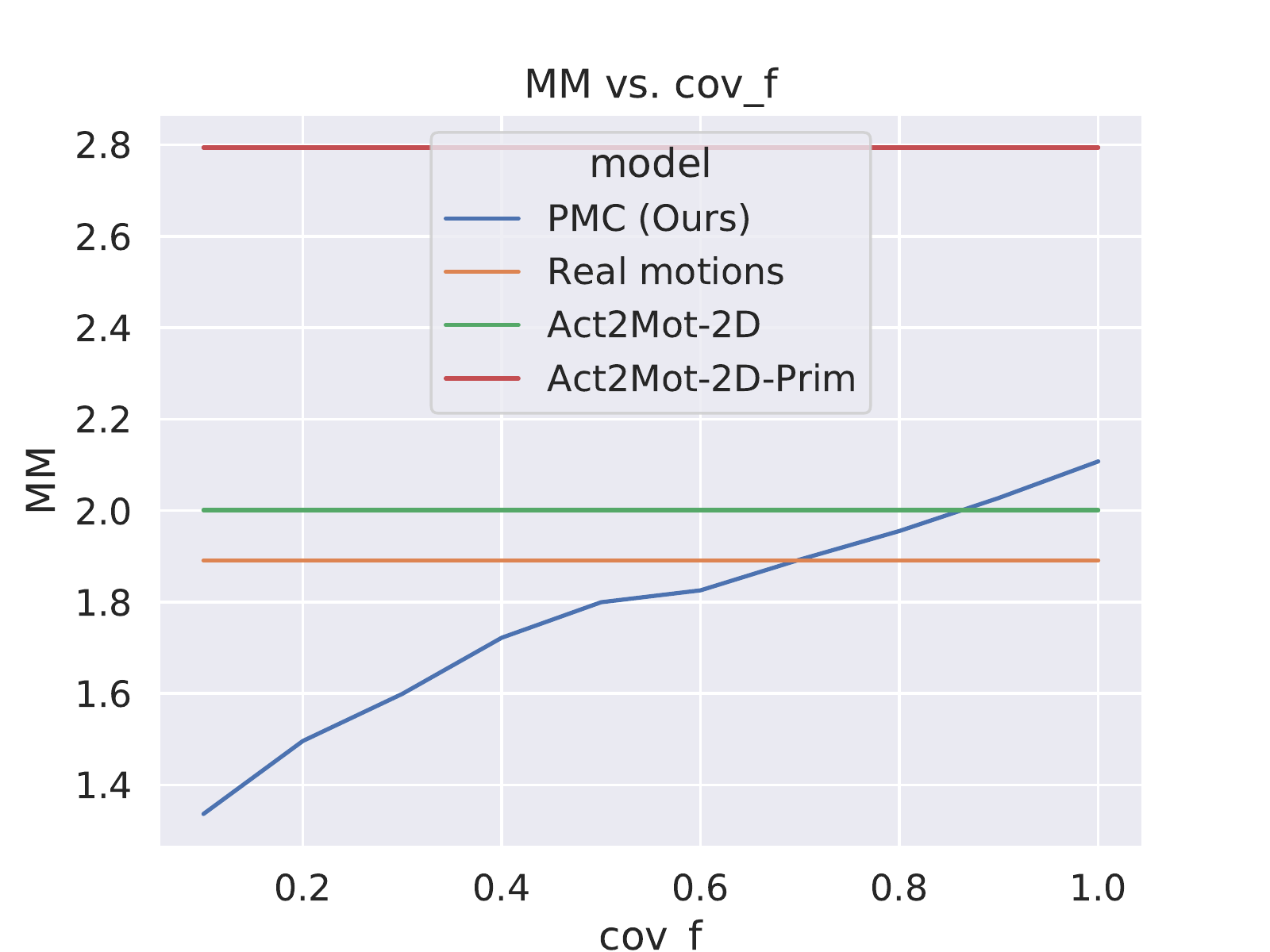}
    \vspace{-2pt}
    \caption{Ablating the covariance factor for evaluating action-conditioned motion synthesis. We compare the \textbf{FID} score, \textbf{Acc} (recognition accuracy), \textbf{Div} (variance across action classes) and \textbf{MM} (variance within action classes).}
    \vspace{-10pt}
    \label{fig:covf-plots}
\end{figure}

We study the effects of varying covariance factor ($\textit{cov\_f}$) on the generated motions. We present the results of our study in Figure~\ref{fig:covf-plots}. We observe that one can get motions with increased diversity and multimodality by increasing the $\textit{cov\_f}$. We observe that higher $\textit{cov\_f}$ factor leads to less satisfactory visual quality, even if quantitatively there is only small drop of the recognition accuracy. Hence, we choose a lower default value of $\textit{cov\_f}$ but this can be modified by the user as needed.

\section{Ablation of Window Size}
\label{sec:ws-ablate}
\begin{figure}[ht]
    \centering
    \begin{center}
    \includegraphics[width=.48\linewidth]{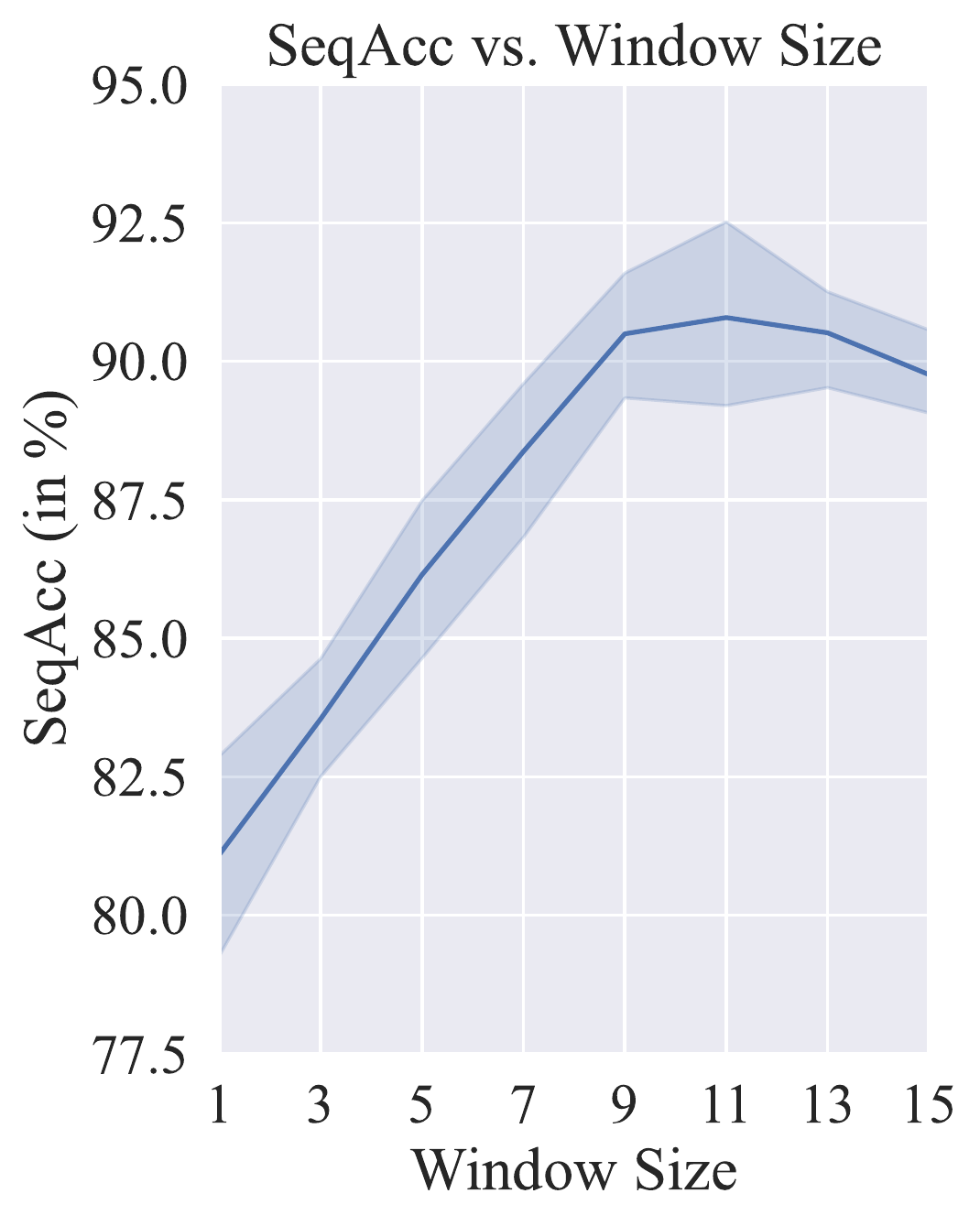}
    \includegraphics[width=.50\linewidth]{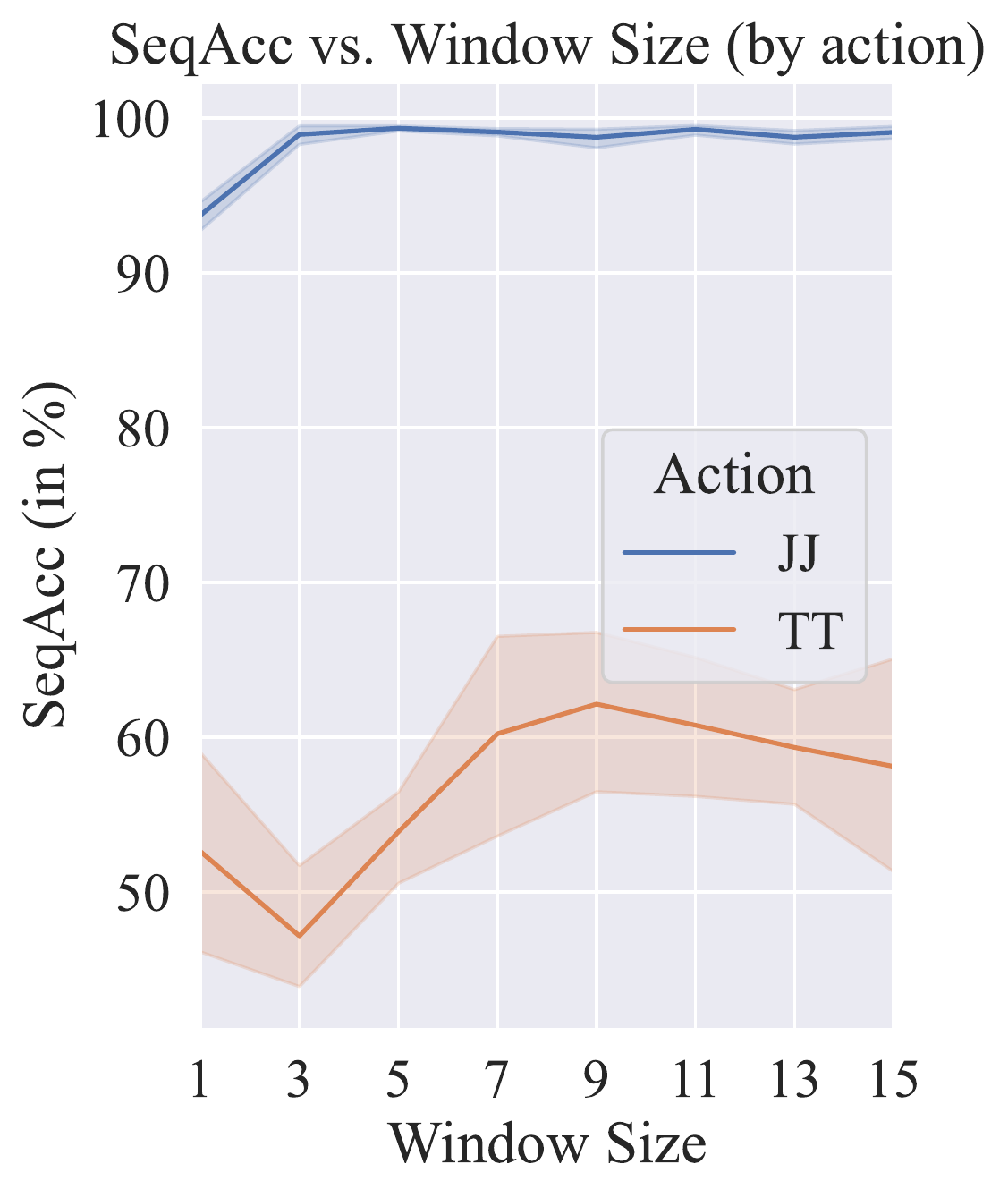}
    \end{center}
    \vspace{-16pt}
    \caption{Ablating the window size for evaluating human motion description (SeqAcc). Left: ablation across all classes. Right: performance breakdown for two classes -- Jumping Jacks (JJ, easy) and Torso Twists (TT, hard).}
    \vspace{-16pt}
    \label{fig:ws-plots}
\end{figure}

We study the effects of varying the window size ($\textit{WS}$) on the motion description accuracy (SeqAcc). We present the results of our study in Figure~\ref{fig:ws-plots}. We observe that across all classes, the performance improves until it plateaus around $\textit{WS}$ equal to 9. We also present a breakdown on two specific classes: Jumping Jacks (JJ) and Torso Twists (TT). We observe that the easier JJ class achieves high accuracy for all values of $\textit{WS}$ while the harder TT class sees improvement with increasing $\textit{WS}$. Intuitively, aggregating temporal context information helps up to certain lengths beyond which it provides no additional help.

\section{Details of the Evaluation Metrics}
\label{sec:metrics}

We use standard metrics used in previous works to evaluate action-conditioned motion synthesis~\cite{guo2020action2motion, petrovich2021action} and controlled motion synthesis from descriptions~\cite{lin2018generating, ahuja2019language2pose, ghosh2021synthesis}. In this section, we provide details of how these metrics were computed for our evaluations.

\subsection{Action-Conditioned Motion Synthesis}
We evaluate action-conditioned motion synthesis on four quantitative metrics that together try to capture the correctness and diversity of the synthesized motions. We compute the FID score, recognition accuracy (Acc), diversity (Div) and multimodality (MM). We train a simple RNN-based action classifier on the single repetition examples collected for each concept class. We use this for computing the recognition accuracy and also as a feature extractor for computing all other metrics. Closely following Guo~\etal~\cite{guo2020action2motion}, we report the average of 20 independent runs for each metric.

\paragraph{FID score:} We extract features of 1000 generated and real motions each (from test set with replacement). We then compute the FID between the generated distribution and real distribution. FID captures the overall quality of the generated motions.

\paragraph{Accuracy (Acc):} We use the RNN-based action classifier to classify 1000 generated motions into classes. This indicates if the generated motions are recognized as the specified classes.

\paragraph{Diversity (Div):}  To capture the variance of generated motion across classes, we generate two subsets $u$ and $v$ of 200 motions each and extract features $\{u_1, u_2, \cdots, u_{200}\}$ and $\{v_1, v_2, \cdots, v_{200}\}$. We then compute the diversity metric defined as $$\textnormal{Div} = \frac{1}{200}\sum_{n=1}^{200} ||u_n - v_n||_2.$$

We use the RNN-based action classifier trained for the ``Accuracy'' measure (without the last linear layer for classification) as the feature extractor.

\paragraph{Multimodality (MM):}  To capture the variance of generated motion within classes, for each motion concept $c$, we generate two subsets $u_c$ and $v_c$ of 20 motions each and extract features $\{u_{c,1}, u_{c,2}, \cdots, u_{c,20}\}$ and $\{v_{c,1}, v_{c,1}, \cdots, v_{c,20}\}$. We then compute the multimodality metric defined as $$\textnormal{MM} = \frac{1}{C \times 20}\sum_{c=1}^{C}\sum_{n=1}^{20} ||u_{c,n} - v_{c,n}||_2$$

We use the RNN-based action classifier trained for the ``Accuracy'' measure (without the last linear layer for classification) as the feature extractor.

\subsection{Controlled Motion Synthesis}
We evaluate controlled motion synthesis on two metrics. For generated pose sequence $P = \{p_1, p_2, \cdots, p_T\}$ and ground truth pose sequence $\bar{P} = \{\bar{p}_1, \bar{p}_2, \cdots, \bar{p}_T\}$, we compute the following two metrics.

\paragraph{Average Positional Error (APE):} We define APE as the $L_2$ distance between the joint keypoints averaged across all joints over the time duration. Mathematically, it is defined as $$\textnormal{APE} = \frac{1}{J \times T}\sum_{t=1}^{T}\|p_t - \bar{p}_t\|_2. $$
Since the generated and ground-truth sequences could be of different lengths, we use Dynamic Time Warping (DTW)~\cite{salvador2007toward} to align these sequences.

\paragraph{Average Variance Error (AVE):} We define AVE as the $L_2$ distance of variances of the generated motion with ground truth motion. We define variance of joint $j$ as in a pose sequence $P = \{p_1, p_2, \cdots, p_T\}$ as: $$\sigma(j) = \frac{1}{T - 1} \sum_{t=1}^{T} \|p^j_t - \mu_p^j\|_2,$$where ${\mu_p^j}$ is average location of joint $j$ across the time duration, \ie, ${\mu_p^j} = \frac{1}{T} \sum_{t=1}^T p^j_t$. Similarly, we can define $\bar{\sigma}(j)$ for the groundtruth pose sequence $\bar{P}$. We then define AVE as: $$\textnormal{AVE} = \frac{1}{J}\sum_{j=1}^{J}\|\sigma(j) - \bar{\sigma}(j)\|_2.$$

\section{Limitations and Societal Impact}
\label{sec:limitations}
Human motion has favorable structural properties such as constraints on the motion of keypoints and repetition that our  methods exploit. Our method relies on primitive extraction which could be difficult in videos with occlusions or noisy pose detections. It is also not immediately clear how well these methods translate to motion of other entities.

Our research has  potential positive societal impacts, with future applications in sports training and assistive technologies in the rehabilitation of disabled persons. On the other hand, like all other visual content generation methods, our method might be exploited by malicious users with potential negative impacts. In our code release, we will explicitly specify allowable uses of our system with appropriate licenses.

{\small
\bibliography{motionprog,bpi}
\bibliographystyle{ieee_fullname}
}